\pdfoutput=1
\documentclass[11pt]{article}

\usepackage[]{EACL2023}
\usepackage{times}
\usepackage{latexsym}
\usepackage{graphics}
\usepackage{graphicx}
\usepackage{subfig}
\usepackage{array}
\usepackage{booktabs}
\usepackage{multirow}
\usepackage{bbm}
\usepackage[normalem]{ulem}
\usepackage{enumitem}
\usepackage{makecell}
\usepackage{xcolor}
\usepackage{tabularx}
\usepackage{arydshln}
\usepackage{adjustbox}

\newcolumntype{C}{>{\centering\arraybackslash}X}

\usepackage[T1]{fontenc}
\usepackage[utf8]{inputenc}

\usepackage{microtype}

\usepackage{xspace}
\newcommand{\method}{\texttt{XLGen}\xspace}

\newcolumntype{P}[1]{>{\centering\arraybackslash}p{#1}}

\title{
Cluster-Guided Label Generation in Extreme Multi-Label Classification
}
\author{
    Taehee Jung  $^\clubsuit$ \thanks{$^*$ Part of this work was done during an internship at Amazon Alexa AI.}
    \quad \quad Joo-Kyung Kim $^\clubsuit$
    \quad \quad Sungjin Lee $^\clubsuit$
    \quad \quad Dongyeop Kang $^\heartsuit$ \\
    \\
  $^\clubsuit$Amazon Alexa AI\quad \quad  $^\heartsuit$University of Minnesota   \\
  {\tt $\{$jungtaeh,jookyk,sungjinl$\}$@amazon.com}
   \quad \texttt{dongyeop@umn.edu}
  }

\begin{document}
\maketitle
\begin{abstract}

%Most existing models for extreme multi-label classification (XMC) use a classification method, training an independent classifier for each label.
For extreme multi-label classification (XMC), existing classification-based models poorly perform for tail labels and often ignore the semantic relations among labels, like treating ``Wikipedia'' and ``Wiki'' as independent and separate labels.
In this paper, we cast XMC as a \textit{generation} task (\method), where we benefit from pre-trained text-to-text models.
However, generating labels from the extremely large label space is challenging without any constraints or guidance.
% \method can be biased to frequent head labels and the order of labels given to the label generator, due to the extremely large number of label space.
We, therefore, propose to guide label generation using label cluster information to hierarchically generate lower-level labels.
%In addition, we find that random label permutation \cite{simig-etal-2022-open} for multi-label supervision does not outperform prior classification baselines for both head and tail labels, not only because it fails to predict labels accurately, but also because it loses label co-occurrence information by shuffling the sequence order. 
% To improve the performance, we propose to 
We also find that frequency-based label ordering and using decoding ensemble methods are critical factors for the improvements in \method.
%\sout{frequency-based label ordering, instead of random permutation \mbox{\cite{simig-etal-2022-open}}, shows noticeable improvements in \method.}
%\jk{frequency-based label ordering and using diverse decoding methods are critical factors for the improvements in \method.}
%and different diversity-promoted sampling methods
% compare its effects along with different decoding order strategies.
%However, generated labels are often noisy, and biased to the frequency and order of labels given to the generative model.
%Simple label permutation \cite{simig-etal-2022-open} does not outperform the prior classification baselines because it still suffers from generating long-tail and unseen but positive labels.
%To address this curse of the extreme label size, we propose to constrain label generation using label clustering information to generate lower-level labels given high-level cluster IDs.
\method with cluster guidance significantly outperforms the classification and generation baselines on tail labels, and also generally improves the overall performance in four popular XMC benchmarks.
In human evaluation, we also find \method generates unseen but plausible labels.
Our code is now available at {\small \url{https://github.com/alexa/xlgen-eacl-2023}}.
%\footnote{Our code will be publicly available upon acceptance.}

%In particular, \method shows a noticeable generalization effect on semantically correlated labels and often generates unseen but positive labels.
%\method will gradually benefit from generative models as pre-trained language models become more powerful and larger.
\end{abstract}

\section{Introduction}
\label{sec:intro}

% For instance, imagine users want to label a blog post with their own tags.
% Each user may have different views and their own ways of tagging the post. 
% This causes an extremely long tail distribution of labels in existing XMC benchmarks, as depicted in Figure \ref{fig:genXMC_freq}.
\begin{figure}[t!]
\centering
{
{\includegraphics[trim=4.2cm 0.2cm 4.2cm  1.3cm,clip,width=1.0\linewidth]{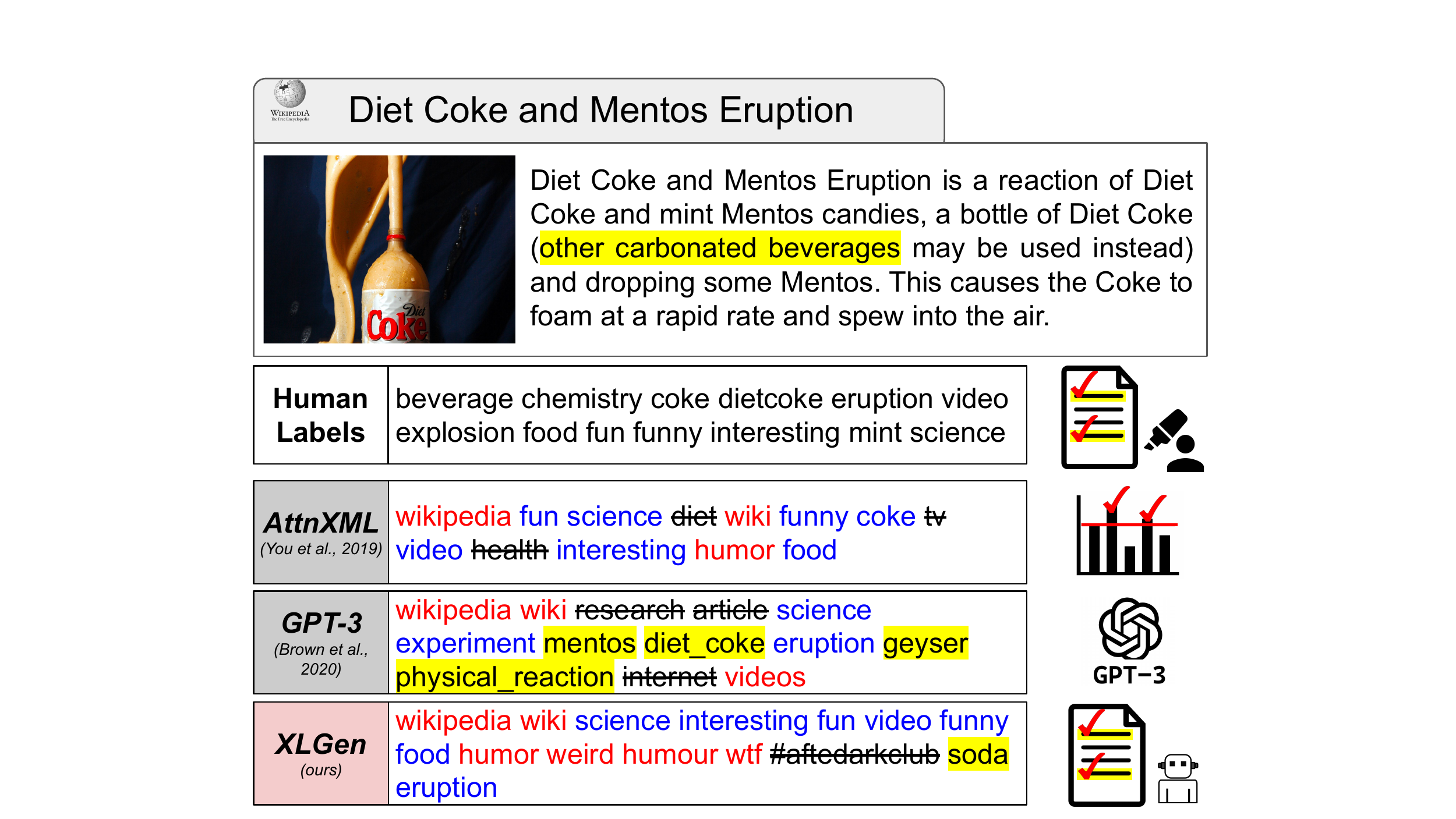}
}
}
\caption{\label{fig:example_fincher} The predicted and generated labels from AttentionXML~\citep{you2019attentionxml}, GPT-3~\citep{brown2020language}, 
 and \method-\texttt{BCL}, respectively, for Wikipedia page on diet coke and mentos eruption.
We marked labels to be correct (\textcolor{blue}{blue}), wrong (\sout{strikethrough}), and positive unlabeled (\setlength{\fboxsep}{0pt}\textcolor{red}{red}).
Our \method could generate completely new labels from input text, e.g., 
% a new label \setlength{\fboxsep}{0pt}\colorbox{orange!90}{fincher} comes from the name of director, and 
\setlength{\fboxsep}{0pt}\colorbox{yellow!90}{soda}, inferred from context that other carbonated beverages can replace diet coke.}
\end{figure}

Extreme multi-label classification (XMC) is a task to predict multiple relevant labels for a given input where the label space is extremely large.
Conventional approaches for XMC decompose the problem into a set of binary classifications, training one-vs-all classifiers for each label.
However, they encounter several issues in practical use cases.

\begin{figure}[t!]
\hspace*{-0.5cm}
\centering
\makebox[1.1\linewidth][c]{
\subfloat[\label{fig:genXMC_freq} Frequency Histogram]{\includegraphics[trim=0cm 0cm 0cm  0cm,clip,width=0.55\linewidth]{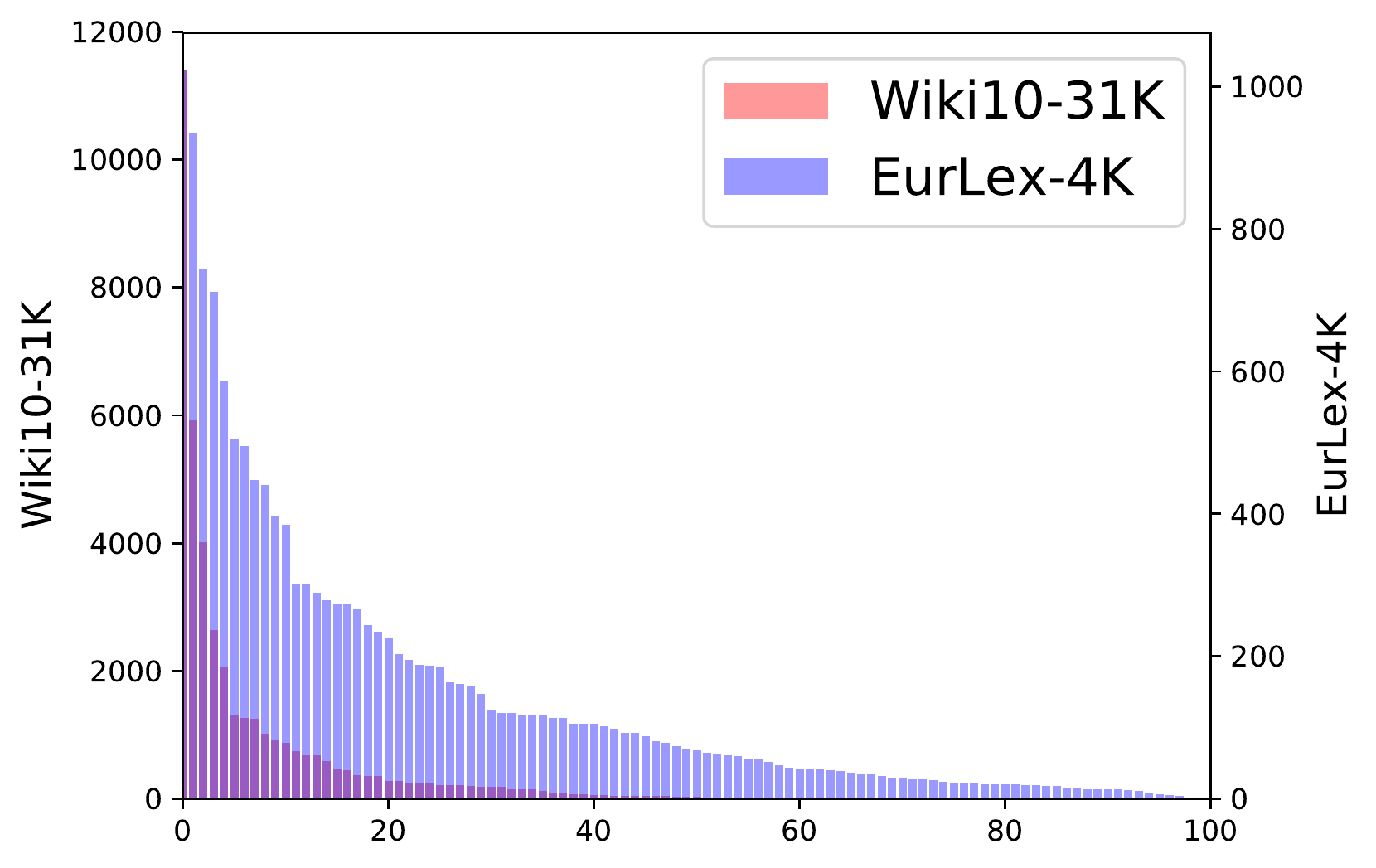}}
\subfloat[\label{fig:label_recall} Label-wise Recall]{\includegraphics[trim=0 0 0 0,clip,width=0.48\linewidth]{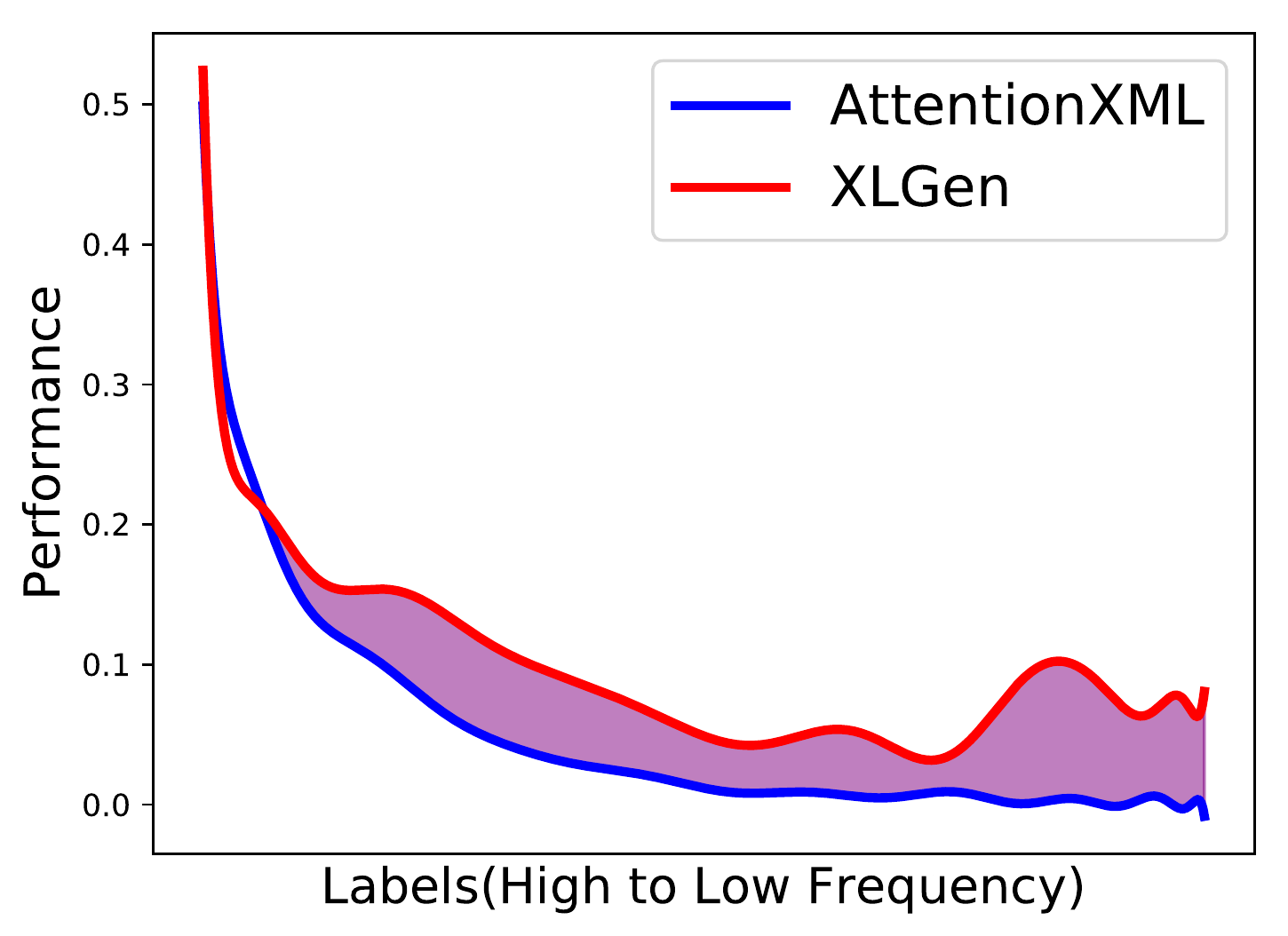}}
}
\caption{
(a) Frequency histograms of top-100 occurring labels in \textsc{EUR-Lex} (blue) and \textsc{Wiki10-31K} (red) \\
(b) Label-wise recall scores from AttentionXML and \method-\texttt{BCL} on Wiki10-31K.
% Labels on the X-axis are sorted by descending frequency order, so the tail labels appear last.
For the presentation, graphs are smoothed by least-squared polynomial regression.
% A purple-shaded area represents the performance gap between AttentionXML and \method-\texttt{BCL}, showing the significant tail label improvement.
}
\end{figure}

First, the labels in XMC are long-tail distributed. 
In other words, only a few labels have sufficient positive samples, thereby the other infrequent labels could be rarely predicted during inference as we see the heavily right-skewed distribution in the long-tail in Figure \ref{fig:genXMC_freq}.
Second, multi-label classification techniques such as one-by-one and label powerset \cite{Gibaja2015ATO} assume independent and identically distributed labels, while the user-generated labels in XMC are dependent on each other.
Moreover, annotated labels are only a portion of possible labels, thus, resulting in positive and unlabeled (PU) setting \cite{Yu2014, Kanehira2016}.
%\sj{you may also want to mention the fact that each user tags only some portion of data}
%For example, in \textsc{Wiki10-31K} dataset, labels like 'science fiction' can be viewed as a sub-category of 'novel', thus, wikipages related science fiction should have novel as a label too.
%\dk{Add more examples of these labels}
%Lastly, the training time and parameters linearly increase with the number of labels.
%Lastly, most of prior works focused on reducing such computational costs via negative samplings for linear one-vs-all training \cite{prabhu2018parabel,jain2019slice,you2019attentionxml,chang2020taming} or via label partitioning with clustering methods~\cite{prabhu2018parabel,wydmuch2018no,yu2020pecos}.

In this paper, we tackle extreme multi-label classification with a \textit{generative approach}, called extreme multi-label generation (\method).
In particular, we fine-tune a pre-trained Transformer-based encoder-decoder model \cite{raffel2019exploring} with input documents and their known positive labels.
This (label) generation approach is more intuitive and closely similar to how humans tag documents with text labels without a fine-grained ontology or guideline.

However, the generated labels from the extremely large label space without any constraints and/or guidance can be noisy and not cover infrequent labels.
To address this issue, we propose a method to leverage label clusters into generation: first, generate cluster IDs of semantically similar labels, and then generate text labels utilizing the cluster IDs as additional contextual inputs.
Specifically, we propose two \method architectures (\method-\texttt{BCL}, \method-\texttt{MCG}) in which such clusters are jointly trained with labels in different ways.
%The cluster information used in training is pre-computed with conventional algorithms like K-Means. 
% Our motivation is that using cluster information is similar to showing label categories to human annotators.
Using clusters for label generation is motivated by showing label categories to human annotators. As an example, humans often start by setting high-level topics first and then hierarchically create actual tags under each high-level topic.
The clusters, however, are treated as additional guidance rather than a constraint since we do not restrict the model to only predict labels under the given clusters.

Similarly to \method, \citet{simig-etal-2022-open} proposed GROOV, which fine-tunes T5 to generate labels in XMC.
% regardless of their orders.
%and focused to show tail label performances.
In particular, GROOV aims to implement a label order invariant training objective by randomly shuffling label orders and using multi-softmax function, which does not penalize if any first tokens of true labels are predicted regardless of the label orders.
However, it does not outperform classification baselines consistently, and we empirically find that label order by frequency helps alleviate the issue in our ablation study.
% because such label invariant objective rather loses the label co-occurrence information.
% We explore various decoding strategies and empirically find that label order frequency helps alleviate the issue.

% \begin{figure}[t!]
% \centering
% {
% {\includegraphics[trim=0 0 0 0,clip,width=0.83\linewidth]{img/label_recall.pdf}
% }
% }
% \caption{\label{fig:label_recall} Label-wise recall scores from AttentionXML and \method-\texttt{BCL} on Wiki10-31K.
% Labels on the X-axis are sorted by descending frequency order, so the tail labels appear last.
% For the presentation, graphs are smoothed by least-squared polynomial regression.
% A purple-shaded area represents the performance gap between AttentionXML and \method-\texttt{BCL}, showing the significant tail label improvement.}
% \end{figure}

Our experiment shows that \method (and its variants) outperforms classification baselines on four XMC benchmarks.
Furthermore, \method with cluster guidance (\method-\texttt{BCL} and -\texttt{MCG}) significantly and consistently outperforms the classification and generation baseline (\method-\texttt{base}) on tail labels, respectively. 
% In particular, our two proposed \method architectures achieve consistent improvement over simple fine-tuning of T5 (\method-\texttt{base}).
The effect on tail labels from \method-\texttt{BCL} is demonstrated in Figure~\ref{fig:label_recall}.

Figure~\ref{fig:example_fincher} shows predicted or generated labels from different models.
A Wikipedia page of diet coke and mentos eruption has true labels such as ``beverage'', ``fun'', and ``eruption''.
We find \method can also generate a new positive label ``soda'' based on the context of ``carbonated beverage" in the input text.
From a human evaluation (S\ref{subsec:human}), we find newly generated labels by \method are highly associated with the input texts, which potentially helps automatically find new labels without manual tagging.
%\method will gradually benefit from generative models as pre-trained language models become more powerful and larger.
% Furthermore, as a prompt-based few-shot learning approach has rapidly emerged with state-of-the-art performance on a variety of NLP problems, 
We also show the generated labels from large language models (LLMs) like GPT-3~\citep{brown2020language}: 
we find that the overall performance of in-context learning is significantly less than \method (See \S\ref{sec:feasibility_gpt3} for details), but LLMs could generate reasonable labels with a few examples as \method does.

% Our work is a first attempt to directly generate text labels by utilizing cluster guidance in XMC; we, therefore, provide extensive ablation tests to find the best setup.

% We believe that switching the framework for XMC from classification to generation offers significant opportunities in terms of performance improvement.

\begin{figure*}[t!]
\centering
\subfloat[\label{fig:genXMC_1} \method-\texttt{base}]
{\includegraphics[trim=0cm 4.5cm 15cm  0cm,clip,width=.31\linewidth]{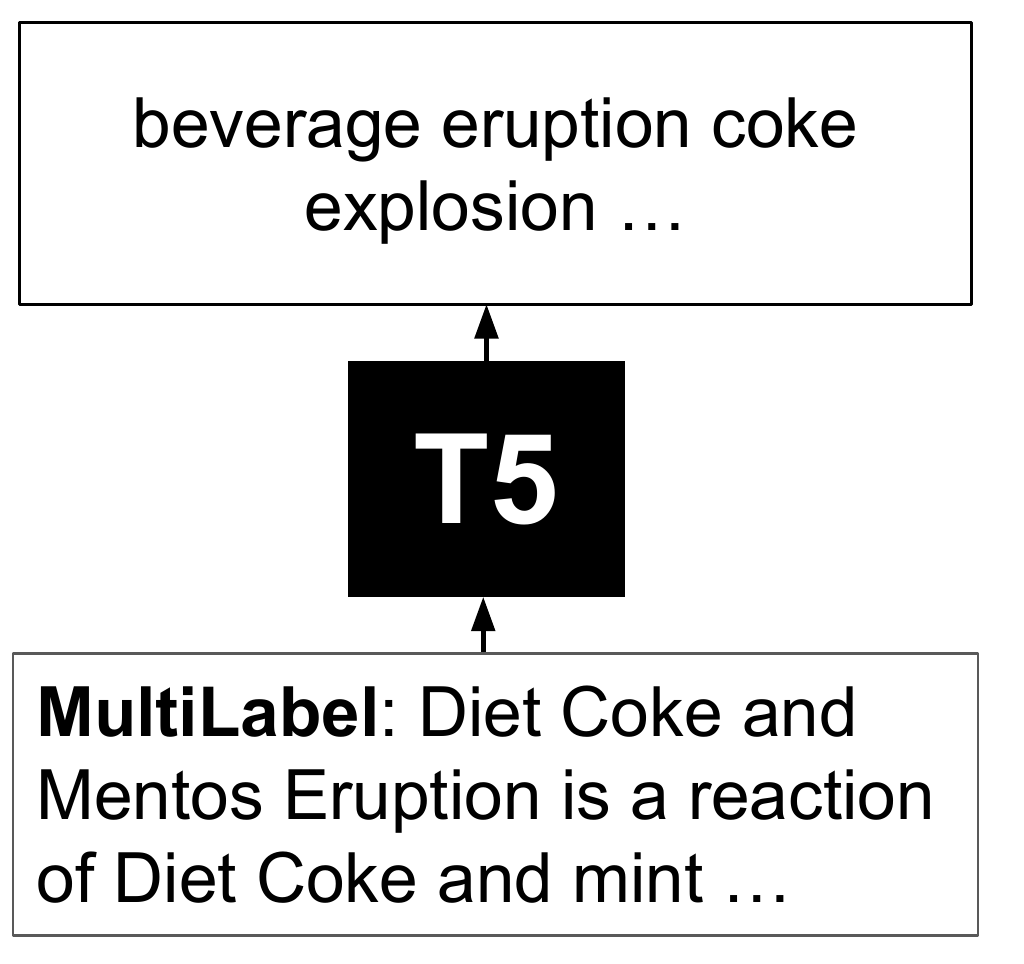}
}
\hspace{0.3 cm}
\subfloat[\label{fig:genXMC_2} \method-\texttt{BCL}]
{\includegraphics[trim=0cm 4.5cm 15cm 0cm,clip,width=.31\linewidth]{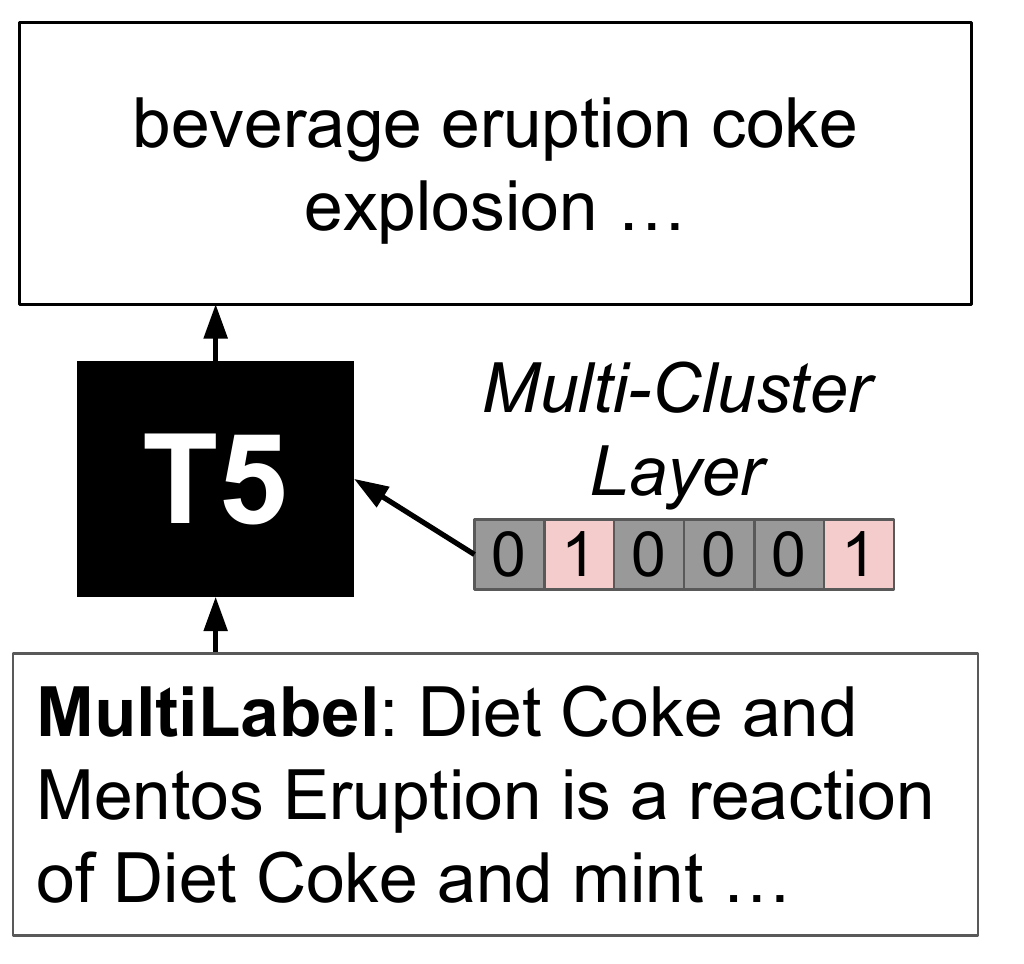}
}
\hspace{0.3 cm}
\subfloat[\label{fig:genXMC_3} \method-\texttt{MCG}]
{\includegraphics[trim=0cm 4.5cm 15cm 0cm,clip,width=.31\linewidth]{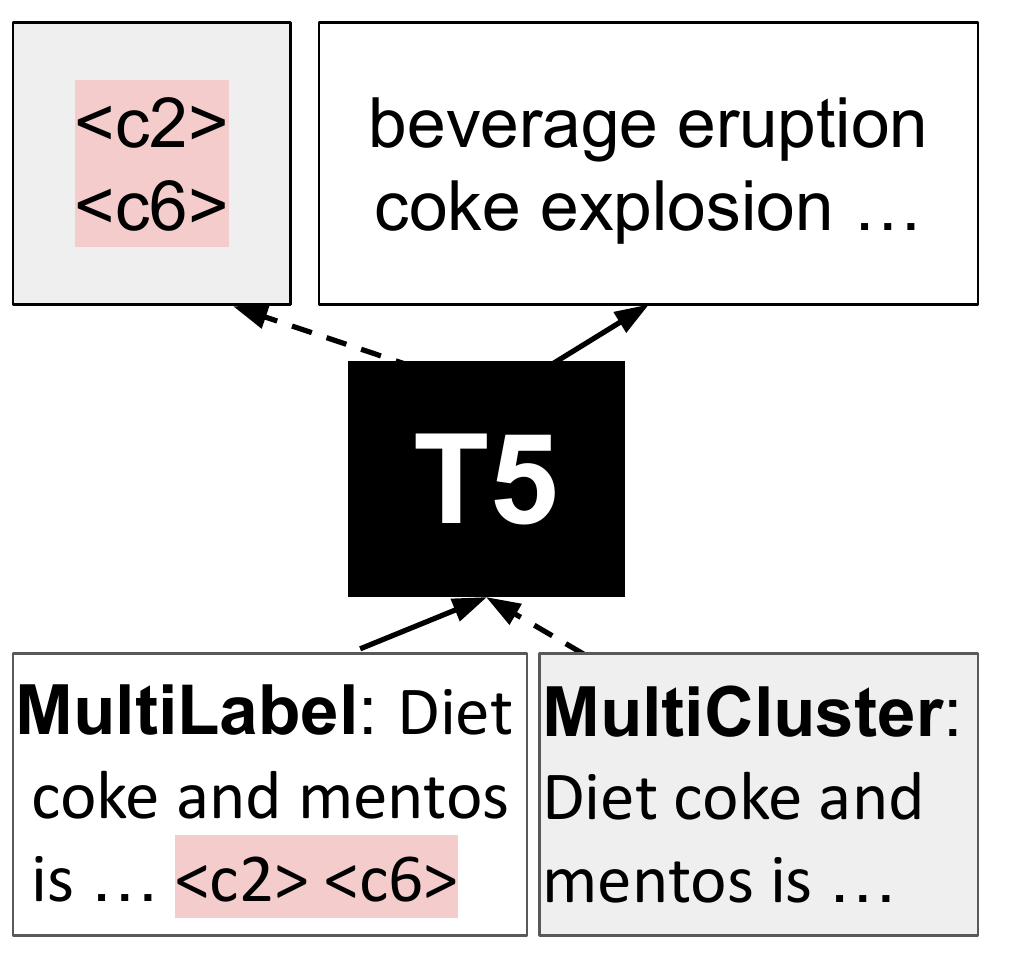}
}
\caption{Three \method architectures, where the basic model can be any pre-trained text-to-text models like T5. 
(a, \texttt{base}): Simple fine-tuning that encodes input text with a prefix of task name and decodes text of label sequences.
(b, \texttt{BCL}): A fine-tuning with a multi-cluster prediction layer as an auxiliary task.
%For inference, such predicted clusters can be incorporated as an input as well (\method-\texttt{BCL}$_{concat}$).
(c, \texttt{MCG}): A multi-task fine-tuning with multi-cluster generation and multi-label generation (MCG); two tasks are trained simultaneously and at decoding time the output of the cluster generation is concatenated to the input for the label generation.}
%\sj{is contained by sounds a bit odd to me.} \sj{it'd great if we could add the clustering stage in the overall architecture as it's not clear in the text.}}
\end{figure*}

\section{Related Work}
\label{sec:related}

% \begin{table}
% \centering
% \begin{adjustbox}{width=0.99\linewidth,center}
% \begin{tabular}{ccc}
% % \hline
% Model &Task  & Clustering Labels\\
% \hline
% AttentionXML & Classification & O  \\
% GROOV & Generation & X  \\
% \method &  Generation & O  \\ 
% \hline
% \end{tabular}
% \end{adjustbox}
% \caption{Comparison of \method with other baselines. \method utilizes cluster information and jointly trains it with label generation in order to capture the label correlations and control the noise in generated labels.
% \vspace{-3mm}}
% \label{tab:model_comparision}
% \end{table}

\paragraph{Extreme multi-label classification (XMC).} 
Classification-based approaches on XMC suffer from dealing with enormous label spaces under the one-vs-all classification setting~\citep{babbar2017dismec,yen2017ppdsparse,jain2019slice}.
To address the efficiency issue, state-of-the-art XMC models partition label space to the scalable subsets via hierarchical clustering~\citep{prabhu2018parabel,wydmuch2018no,you2019attentionxml,chang2020taming,yu2020pecos,tagami2017annexml}, graph-based approximations~\citep{jain2019slice,9839555}, or random forest~\citep{siblini2018craftml}.
However, they still suffer from predicting tail or unseen labels. 
To efficiently deal with such long-tail issues, few-shot learning frameworks and methods~\citep{Gupta21,xiong-etal-2022-extreme} are proposed.
Rather, we show how encoder-decoder language model can improve tail label scores by fine-tuning it with guidance from label clusters. 

\paragraph{LMs and Generative approach in XMC.}
For XMC, pre-trained LMs such as XLNet~\citep{ye2020pretrained} and Transformer~\citep{chang2020taming} are used but only for encoding input texts, thus, it still relies on the classification approach for label prediction.
Previously, other works address multi-label classification with generative approaches~\citep{nam2017maximizing,tsai2020order,yang2018sgm, yang2019deep, zhang2021enhancing} but in much smaller label spaces.
Recently, \citet{simig-etal-2022-open} also used T5 \citep{raffel2019exploring} for directly generating labels in end-to-end manners for XMC, but its performance was not convincing compared to classification-based models.

\paragraph{Positive and unlabeled data.}
In practice, XMC is inherently with positive and unlabeled (PU) setting as the label space is extremely large and it is infeasible to manually review all the labels \citep{JKKim2020}.
Multi-label performances on PU tasks can be simulated by leaving only a few labels per train instance (e.g., leaving 8 out of 10 positive labels in one instance for label deficit rate 20\%) positive~\citep{hu2021predictive}.
% or leaving only a few labels (e.g., 50 out of 100 possible labels for 50\% PU ratio) positive~\citep{kanehira2016multi} 
In this work, we show how \method works on such PU settings in \S\ref{subsec:pu}.

% \textbf{Label semantics in text classification.} 
% For text classification tasks such as named entity recognition and emotional classification, previous works incorporate label semantics into the text embeddings using neural encoders \citep{10.1609/aaai.v33i01.33016642,gaonkar2020modeling,ma-etal-2022-label}.
% \citet{mueller-etal-2022-label} propose a secondary pre-trained T5 model for intent classification optimized by predicting masked intent from input sequences where documents and their corresponding intents are concatenated.
% Naturally, a generative approach like \method also implicitly employs label semantics since it predicts label tokens based on the generated label embeddings in decoder.
% Later, we visualize the label semantics optimized by \method in \S\ref{subsec:lab_semantic}.
\section{Extreme Multi-label Generation (\method) with Cluster Guidance}
\label{sec:method}

Classifying a document with multiple labels can be regarded as tagging a document with possible topical labels, which is basically decoding the free-formed text labels in an encoder-decoder setting.
Moreover, if encoder and decoder are trained on large text corpora, labels are generated with an understanding of their lexical variations and semantic similarities. %between labels.
%\sout{Since our work is the first attempt to use the pre-trained encoder-decoder language models on the multi-label classification tasks, we incrementally develop from simple to more advanced model architectures to tackle the challenging issues we faced in the development.
%We propose three variants of \method as follows:}
Our baseline framework fine-tunes a pre-trained Transformer using the input text as an encoder input and the label sequences as a decoder output.
%but labels are sorted by frequency order and a softmax function is used as an objective.
In addition, to more effectively address a huge number of labels in a long-tail distribution, we propose two different architectures,
\method-\texttt{BCL} (\S\ref{subsec:finetuning_cluster_prediction}) and \method-\texttt{MCG} (\S\ref{subsec:finetuning_cluster_decoding}), generating labels guided by pre-computed cluster information, inspired by class-based language models~\citep{brown1992class}.
%to leverage the label clusters

\begin{figure}[t!]
\centering
{
{\includegraphics[trim=0cm 3.6cm 6.8cm  0cm,clip,width=1.05\linewidth]{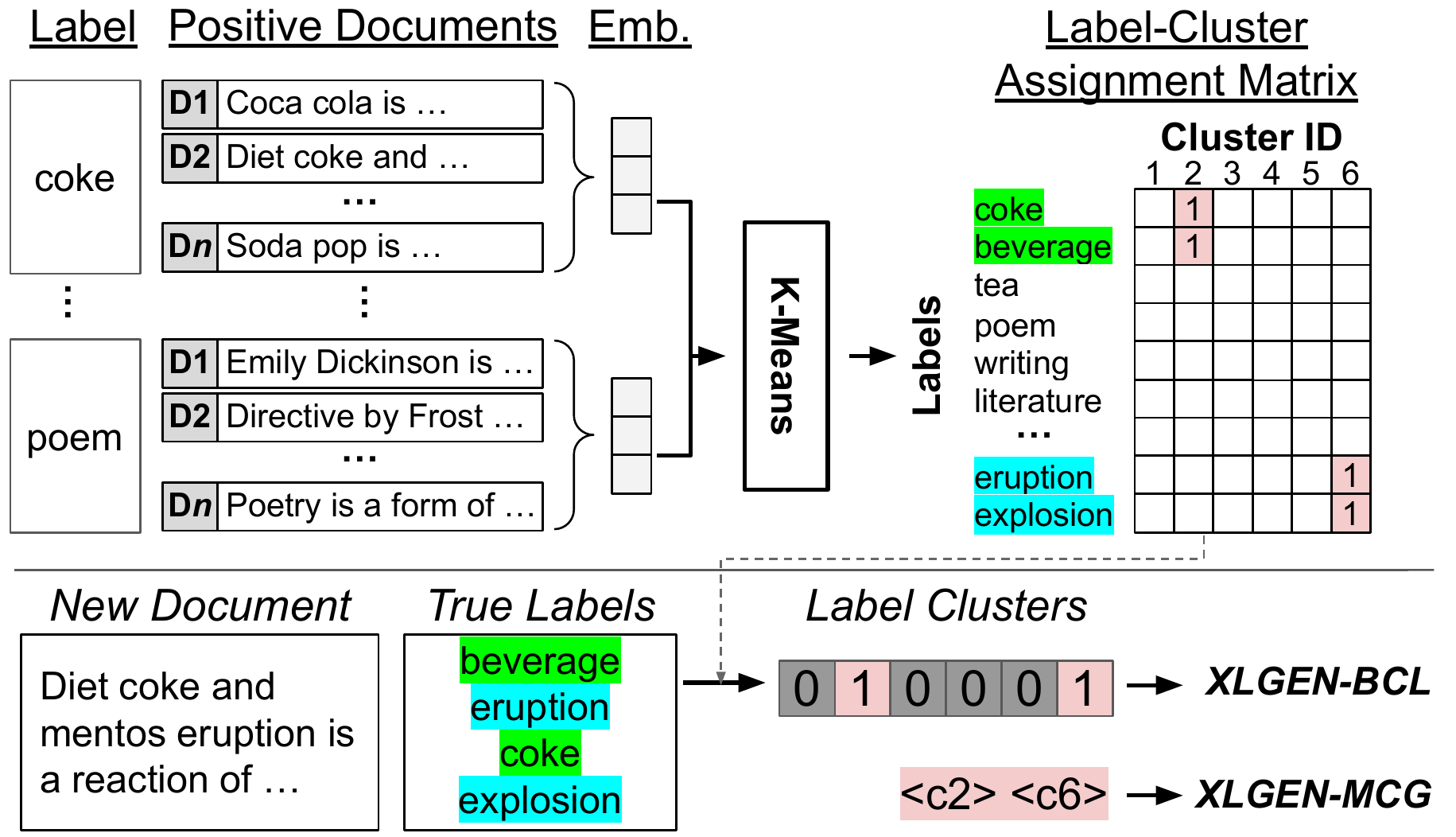}
}
}
\caption{\label{fig:genXMC_cluster} Architecture of pre-computed label clusters.
For each label, we use the averaged embedding of positive documents including the label in train set and compute the label cluster assignment matrix (top).
For training, we assign label clusters for each training document based on the true labels using the label cluster assignment matrix (bottom).}
\end{figure}

\paragraph{Pre-computed clustering.}
We compute label clusters using K-Means algorithm, as depicted in Figure~\ref{fig:genXMC_cluster}.
We first obtain label features using average embedding of positive documents in a train set, following \citet{chang2020taming}, and compute the label-cluster assignment matrix.
Label clusters are assigned to each training document using this matrix and ground-truth labels, and used as a multi-cluster prediction layer for \method-\texttt{BCL} training or sequence of cluster IDs for \method-\texttt{MCG} training.

\subsection{Baseline Fine-Tuning}
%We use a pre-trained text-to-text Transformers such as T5 \cite{raffel2019exploring} and BART \cite{lewis2020bart} as our encoder-decoder framework.
Figure~\ref{fig:genXMC_1} shows our baseline \method which simply fine-tunes text-to-text Transformers, e.g., T5 \cite{raffel2019exploring} or BART \cite{lewis2020bart}, as our encoder-decoder framework on XMC dataset.
\begin{itemize}[noitemsep,topsep=0pt,leftmargin=*]
    \item \textit{Input}: \textbf{task prefix}: input text
    \item \textit{Output}: A sequence of label texts
\end{itemize}
For encoding, we add a prefix token `MultiLabel' to the input text to inform the task type.
Then, the output labels are generated as a sequence of labels in decoding.
The model is fine-tuned with cross-entropy loss ($\mathcal{L}_{xent}$) given the sequence of label texts.
In practice, the order of labels in decoding significantly influences the model performance. 
Following \citet{yang2018sgm}, we sort the target labels in decreasing order of the frequencies.
We also investigate various ordering effects and their impact on performance in \S\ref{subsection:anal_label_ordering}.
% Here, we find that ordering labels by frequencies in training data from high to low achieves the best scores.

% \begin{table*}
% \centering
% \begin{tabular}{lccccccc}
% \hline
% &$|D_{trn}|$ & $|D_{tst}|$ & $T_{avg}$ & $W_{avg}$  & $L$ & $L_{avg}$ &$D_{avg}$\\
% \hline
% \textsc{EurLex-4K} & 15,449 & 3,865 &2,340 &1,241 & 3,956 & 5.30 & 20.79 \\
% \textsc{AmazonCat-13K} & 1,186,239 & 306,782 & 324 & 219 &13,330 & 5.04 & 448.57 \\
% \textsc{Wiki10-31K} &  14,146 & 6,616 & 3,134 & 2,108 & 30,938 & 18.64 & 8.52 \\ 
% \textsc{Wiki-500K} &1,779,881 &769,421 & 1,298 & 917 & 501,070 &4.75 &16.86  \\
% \hline
% \end{tabular}
% \caption{Data statistics of the benchmark datasets; the number of train examples ($|D_{trn}|$), number of test examples ($|D_{tst}|$), averaged number of tokens ($T_{avg}$), averaged number of words ($W_{avg}$), total number of labels ($L$), averaged number of labels per example ($L_{avg}$), and averaged number of examples per label ($D_{avg}$).}
% \label{tab:dataset}
% \end{table*}

\subsection{Fine-Tuning with Cluster Prediction}
\label{subsec:finetuning_cluster_prediction}
% In a generative model, possibly any tokens existing in a vocabulary can be outputted, that is, we expect to generate totally new labels which are potentially correct but not existing in the current label space.
% However, due to this flexibility of the generative model, the output labels could be unconstrained and sometimes noisy. 
% To address this issue, the following two proposed methods, \method-\texttt{BCL} (\S\ref{subsec:finetuning_cluster_prediction}) and \method-\texttt{MCG} (\S\ref{subsec:finetuning_cluster_decoding}), generate labels, being guided by pre-computed cluster information.
%\sj{this sentence doesn't sound natural.} %under a certain grouping constraint. 
% We add 
% the cluster information of target labels as additional input to the decoder. %\sj{reaching this point, I feel strongly that we should have a separate section on clustering before 3.1. Without that, readers would feel uncomfortable by having to imagine the details of clustering for example features used, whether it's used in runtime or not, what a cluster looks like etc.}
% which is trained using given text input and label text information.
Figure~\ref{fig:genXMC_2} shows the fine-tuning of text-to-text with an additional multi-cluster prediction layer (\method-\texttt{BCL}). 
By doing so, we expect the model learns label similarities and hence biases itself to generate labels relevant to the given cluster.
\begin{itemize}[noitemsep,topsep=0pt,leftmargin=*]
    \item \textit{Input}: \textbf{task prefix}: input text
    \item \textit{Multi-Cluster Layer}: a vector of $v_{1},...v_{k}$ where $v_{i}=1$ if $i^{th}$ cluster $c_{i}$ is a positive cluster; otherwise $v_{i}=0$ (1 1 0 ... 1...)
    %\item \textit{Multi-Cluster Layer}:  c$_1$: 1, c$_2$: 1, .. c$_{k}$: 0
    \item \textit{Output}: A sequence of label texts
\end{itemize}

% Here, a multi-cluster layer contains multiple $1$s as much as the number of labels at max.
%Label clusters are pre-computed using existing cluster with label feature generated by input text. 
%See \S\ref{subsec:anal_cluster} for more details.
%\sout{$k$-means clustering}
The multi-cluster prediction layer is a vector of 0 or 1 that corresponds to the assigned clusters of instance, and is trained using the sequence of the last layer's hidden states of the encoder with a binary cross-entropy loss, $\mathcal{L}_{bce}$.
The final objective is as follows:
\begin{equation}
   \mathcal{L}_{xmc-bcl} =  \mathcal{L}_{xent} + \lambda \mathcal{L}_{bce}
\end{equation}
where $\mathcal{L}_{xent}$ is a cross-entropy loss term for the original text-to-text framework and $\mathcal{L}_{bce}$ is a binary cross entropy loss term for the cluster layer.
$\lambda$ is a weighting parameter for controlling $\mathcal{L}_{bce}$, to be chosen by dev-set performance.

% In our experiment, we also experiment with an advanced model that includes pre-computed (or predicted at inference time) cluster indices on the input text, called \method-\texttt{BCL}$_{concat}$.
% \begin{itemize}[noitemsep,topsep=0pt,leftmargin=*]
%     \item \textit{Binary Clustering}: c$_0$: 0, c$_1$: 1, .. c$_{k-1}$: 0
%     \item \textit{Input}: \textbf{task prefix}: input text ; positive cluster indices ({\color{blue}{c$_1$ c$_7$ c$_{11}$..}})    
%     \item \textit{Output}: A sequence of label texts
% \end{itemize}
% Note that at inference time, the positive cluster indices are predicted from the binary cluster layer and added to the input text.\sj{it's not clear how positive cluster indices are different from Binary Clustering. Hence, it's not clear why the latter is a more advanced than the former.}

\subsection{Fine-Tuning with Cluster Decoding}
\label{subsec:finetuning_cluster_decoding}
%\sout{Instead of having an additional objective for clustering, we can fine-tune the text-to-text model to do both clustering and label generation where generation output is constrained on the cluster output.}
\method-\texttt{BCL} utilizes a cluster prediction only as an auxiliary task to improve representations for a label prediction, thus, predicted clusters are not used in inference.
Figure~\ref{fig:genXMC_3} shows the third variant, \method with a multi-cluster generation (\texttt{MCG}), which leverages predicted clusters as additional input tokens so that the cluster information can be used in inference.

\begin{itemize}[noitemsep,topsep=0pt,leftmargin=*]
    \item \textit{Input1}: \textbf{task prefix}: input text ; a sequence of positive cluster IDs ({{c$_1$ c$_2$ c$_{11}$..}})    
    \item \textit{Output1}: A sequence of label texts
    \item \textit{Input2}: \textbf{cluster prediction prefix}: input text
    \item \textit{Output2}: A sequence of positive cluster IDs ({{c$_1$ c$_2$ c$_{11}$..}})
\end{itemize}

We add a sequence of cluster IDs to the input text so that cluster information can be used while training (Input1-Output1).
On the other hand, we have a new task with a clustering prefix `MultiCluster' appended to the input text and predicts the sequence of labels (Input2-Output2). 
In training, these two tasks are trained simultaneously.
Note that in inference, the predicted cluster IDs are appended to Input1 for the final label generation.
% This architecture takes advantage of text-to-text models, which is originally designed for multi-task learning. \sj{can we clarify why it's a good thing or interesting to take advantage of T2T models?}
\section{Experiments}
\label{sec:experiment}

\begin{table}[t]
\centering
\begin{adjustbox}{width=\linewidth,center}
\begin{tabular}{@{}lcccc@{}}
\hline
&$|D_{trn}|$ & $|D_{tst}|$ & $|L_{seen}|$  & $|L_{unseen}|$ \\
\hline
\textsc{EurLex-4K} & 15,449 & 3,865 & 2,473 & 155 \\
\textsc{AmznCat-13K} & 1,186,239 & 306,782 & 13,275 & 0 \\
\textsc{Wiki10-31K} &  14,146 & 6,616 & 21,060 & 991  \\ 
\textsc{Wiki-500K} &1,779,881 &769,421 & 498,152 & 917 \\
\hline
\end{tabular}
\end{adjustbox}
\caption{Data statistics of the benchmark datasets; the number of train examples ($|D_{trn}|$), number of test examples ($|D_{tst}|$), number of labels in both train and test set ($|L_{seen}|$), and number of labels only in test set ($|L_{unseen}|$), which is zero-occurred labels in Table~\ref{tab:performance_fewshot}.}
\label{tab:dataset}
\end{table}

\subsection{Experimental Setups}

\begin{table*}[t!]
\centering
% \small
\begin{tabular}{@{}P{2.6cm}
P{0.8cm}P{0.8cm}P{0.8cm}P{0.8cm}P{0.8cm}P{0.8cm}P{0.8cm}P{0.8cm}}
\hline
&\multicolumn{2}{c|}{{\textsc{EurLex-4K}}}
&\multicolumn{2}{c|}{{\textsc{AmznCat-13K}}}
&\multicolumn{2}{c|}{{\textsc{Wiki10-31K}}}
&\multicolumn{2}{c}{{\textsc{Wiki-500K}}}\\
\cline{2-9}
&$Mic.$ & $Mac.$ 
&$Mic.$ & $Mac.$ 
&$Mic.$ & $Mac.$
&$Mic.$ & $Mac.$ \\
\hline
XR-Transformer & 39.1&12.3 & 64.0&17.0 & 21.4&2.8 & 30.5&7.8  \\
XR-Linear & 44.6&15.1 & 53.2&18.6 & 19.2&3.6 & 17.2&3.3  \\
AttentionXML & 59.9&24.9 &\underline{70.1}&30.0 & 37.3&4.6 & 53.6&21.0   \\
\hline
\method-\texttt{base} & 59.8&27.5 & 69.8&\underline{38.8} & \textbf{37.6}&\textbf{9.9} & \underline{55.1}&\textbf{35.0}  \\ 
\method-\texttt{BCL} & \textbf{60.7}&\textbf{28.4} & 70.0&37.7 & \textbf{37.6}&\underline{9.8} & \textbf{55.4}&33.5\\
\method-\texttt{MCG} & \underline{60.2}&\underline{28.2} & \textbf{71.8}&\textbf{46.4} & \underline{37.4}&9.6  & \textbf{55.4}&\underline{33.6} \\
\hline
\end{tabular}
%\end{adjustbox}
\caption{Full label performance. 
We report micro-averaged ($Mic.$) and macro-averaged ($Mac.$) F1 scores.
%The best scores are \textbf{bold} and the second best scores are \underline{underlined}.
}
\label{tab:performance_full}
\end{table*}

\paragraph{Datasets.}
We use four widely used XMC benchmark datasets; three large-scale datasets with 4K$\sim$30K labels (\textsc{EurLex-4K},\textsc{AmznCat-13K}, and \textsc{Wiki10-31K}) and one very-large-scale datasets with 500K labels (\textsc{Wiki-500K}).
See Table~\ref{tab:dataset} for the detailed data statistics.
%\footnote{We do not use Amazon-670K and Amazon-3M which were evaluated in \citet{yu2020pecos} because their labels are item IDs, which are difficult to benefit from a pre-trained decoder.}

\paragraph{Baseline and \method Training.}
We compare \method with three state-of-the-art baselines in XMC tasks;  AttentionXML~\citep{you2019attentionxml}, X-Transformer~\citep{chang2020taming}, and  XR-Linear~\citep{yu2020pecos}.
Note that all baseline models partition labels using hierarchical clustering.
See \ref{appendix:baseline} for the detailed setups of baselines. Note it is common to upscale scores by ensemble learning for XMC baselines. However, for a fair comparison, we do not use any ensemble models for \method and baselines.

We train \method with T5 because BART \cite{lewis2020bart} performs worse for our task as shown in Table \ref{tab:performance_bart}.
%, one of the most popular text-to-text Transformer models.
By default, we sort labels in the decreasing frequency order to provide the training target sequence and infer the labels by beam search with size 5.
For \method-\texttt{BCL} and \method-\texttt{MCG}, cluster sizes are optimized by dev set performance.
We get the input text embedding by averaging the last hidden states from the pre-trained T5 encoder since T5 model does not have a CLS token.
See~\ref{appendix:hyperparam} to check more details.

\begin{table}[t!]
\begin{adjustbox}{width=1.0\linewidth,center}
\begin{tabular}{@{}P{2.6cm}
P{0.6cm}@{\hskip 2mm}P{0.6cm}P{0.6cm}@{\hskip 2mm}P{0.6cm}P{0.6cm}@{\hskip 2mm}P{0.6cm}}
\hline
\multirow{2}{*}{}&\multicolumn{2}{c|}{\small{\textsc{EurLex-4K}}}
%&\multicolumn{2}{c|}{\small{\textsc{AmznCat-13K}}}
&\multicolumn{2}{c|}{\small{\textsc{Wiki10-31K}}}
&\multicolumn{2}{c}{\small{\textsc{Wiki-500K}}}\\
\cline{2-7}
&{0-st} &{1-st}
%&{0-st} &{1-st}
&{0-st} &{1-st} 
&{0-st} &{1-st} \\
\hline
XR-Transformer & 0.0 & 0.5 
%& - &  0.0 
& 0.0 & 1.7 
 & 0.0 & 0.0 \\
XR-Linear & 0.0 & 1.1 
%&  - &  0.0 
& 0.0 & 2.3 & 0.0 & 0.1 \\
AttentionXML & 0.0 & 2.4 
%& - &  0.0 
& 0.0 & 0.2 & 0.0 & 1.3 \\
\hline
\method-\texttt{base}  & 3.2 & \underline{3.5} 
%& - & \textbf{9.0} 
& 2.9 & \textbf{8.4}
 &22.5 & 24.1\\ 
\method-\texttt{BCL} &\underline{4.3} &\textbf{4.1}  %& - & \underline{7.9}  
&\underline{3.3} &\textbf{8.4}  &\underline{23.2} &\underline{24.8} \\
\method-\texttt{MCG} &\textbf{4.5} & 2.7 
%& - & 6.5 
&\textbf{11.1} &\underline{8.1}
 &\textbf{23.7} &\textbf{25.5}  \\ 
\hline
\end{tabular}
\end{adjustbox}
\caption{Macro-averaged F1 scores in tail labels, which never occurred (0-st) or occurred once (1-st) in train set.
% Note that \textsc{AmznCat-13K} does not have any zero occurred labels in test set, thus we mark it as \textbf{-}.
}
\label{tab:performance_fewshot}
\end{table}

\subsection{Evaluation Metrics}
Following the prior work in XMC, we report F1 score (F@k) of top-k label probabilities as a supplementary metric in \ref{appendix:supp_score}.
However, such ranking metrics are not applicable to label generation tasks since the generative model only output positive label texts \textit{sequentially} and the order of generated labels does not align with the confidence of the label; in other words, the formerly generated labels do not need to be more confident than the latter ones.
Thus, we use conventional multi-label classification metrics, like micro-averaged F1 score ($Mic.$) and macro-averaged F1 score ($Mac.$), as main evaluation metrics.

In principle, evaluating XMC task with the ranking format is not appropriate for most cases as it requires predicting the number of correct labels as well~\citep{amigo-delgado-2022-evaluating}.
%For baseline models, we first consider following~\citet{nam2017maximizing}'s way to choose predicted labels where the predicted probability is greater than 0.5 (threshold).
%However, as two baselines (XR-Transformer, XR-Linear) do not use cross-entropy objective and rather stick to the ranking-based objective (e.g., hinge-loss),
We therefore select predicted labels only when the predicted score is greater than the threshold optimized from the validation set as in \citet{you2019attentionxml}.

\begin{table}[t!]
\centering
\begin{adjustbox}{width=1.0\linewidth,center}
\begin{tabular}{@{}P{2.6cm}
P{1.8cm}P{1.8cm}P{1.6cm}}
\hline
&\small{\textsc{EurLex-4K}}
&\small{\textsc{Wiki10-31K}}&\small{\textsc{Wiki-500K}} \\ \\
\hline
XR-Transformer 
& 8.6 
%& 13.3
& 3.3
& 7.0 \\
XR-Linear 
& 8.8 
%& 11.8
& 2.7
& 2.8 \\ 
AttentionXML 
& 18.7 
%& 14.6 
& 1.3
& 11.3 
\\
\hline
\method-\texttt{base}
& 18.8 
%& \textbf{41.8} 
& 7.7 
& \underline{31.6}\\ 
\method-\texttt{BCL} 
& \underline{19.3} 
%& \underline{41.1}
& \underline{8.0}
& 31.4 \\
\method-\texttt{MCG} 
& \textbf{21.2}
%& 38.3 
& \textbf{10.1} 
& \textbf{32.7} \\
\hline
\end{tabular}
\end{adjustbox}
\caption{Macro-averaged F1 scores in PU setting (50\% of label deficit ratio).}
\label{tab:performance_pu}
\end{table}

\subsection{Results}
We compare performances of \method and baselines in full labels (Table~\ref{tab:performance_full}), tail labels (Table~\ref{tab:performance_fewshot}), and PU data setting (Table~\ref{tab:performance_pu}).
For tail label and PU setting, we do not include \textsc{AmaznCat-13K} as it does not have zero-occurred labels.
%We use conventional micro-averaged ($Mic.$) and macro-averaged ($Mac.$) F1 scores as main metrics.
The best scores are \textbf{bold} and the second best scores are \underline{underlined}.
See~\ref{appendix:supp_score} for the full scores on tail labels and PU setting.

\paragraph{Full label performance.}\label{subsec:result_full}
In the evaluation with full benchmark sets, all the \method models show outperforming or competitive performance compared to the classification-based baselines. 
For macro F1 scores, \method models hugely outperform the baselines, which empirically represents that our approach is strong at predicting infrequent but correct labels. 
In other words, \method models are less biased to predicting frequent labels. 
Compared to \method-\texttt{base}, both \method-\texttt{BCL} and \method-\texttt{MCG} generally show better performance, which demonstrates the effectiveness of the cluster prediction as an auxiliary loss. 

\paragraph{Tail label performance.}\label{subsec:fewshot}
%We expect that \method predicts long-tail labels more correctly than existing classification-based baselines, so we report few-shot label scores. 
%Moreover, as the label space by \method is not limited, we even expect to generate unseen but correct (or incorrect) labels.
%The ranking-based metrics like Precision@k cannot capture the model performance on tail labels. 
We measure macro F1 scores only for tail labels which never or one-time occur in the train set.
% We analyze few-shot label scores to show how \method works in long-tail labels. 
% Here we measure micro and macro F1 scores only for extremely long-tail labels which never (zero-shot) or one time (one-shot) occur in train dataset, called few-shot scores.
%Thus we measure performances only for long-tail labels which never (zero-shot), one time (one-shot), or less than 5 times occur (5-shot) in train dataset which is called few-shot scores.
We find every baseline extremely suffers from the tail labels, while \method shows significant improvements, demonstrating the power of generative models for long-tail labels.
%(See few-shot columns in Table~\ref{tab:performance_total}).
% In addition, score gap between \method and other baselines get larger in such few-shot label settings.
Surprisingly, \method even predicts never-seen, zero-occurred labels, only inferred from the semantic meaning of the input text.
Similarly to full label performance, \method-\texttt{BCL} and \method-\texttt{MCG} perform better than \method-\texttt{base}, indicating that guidance of label cluster improves tail label performance as well.  

\begin{figure}[t!]
\centering
{
\includegraphics[trim=0cm 0cm 0cm  0cm,clip,width=.9\linewidth]{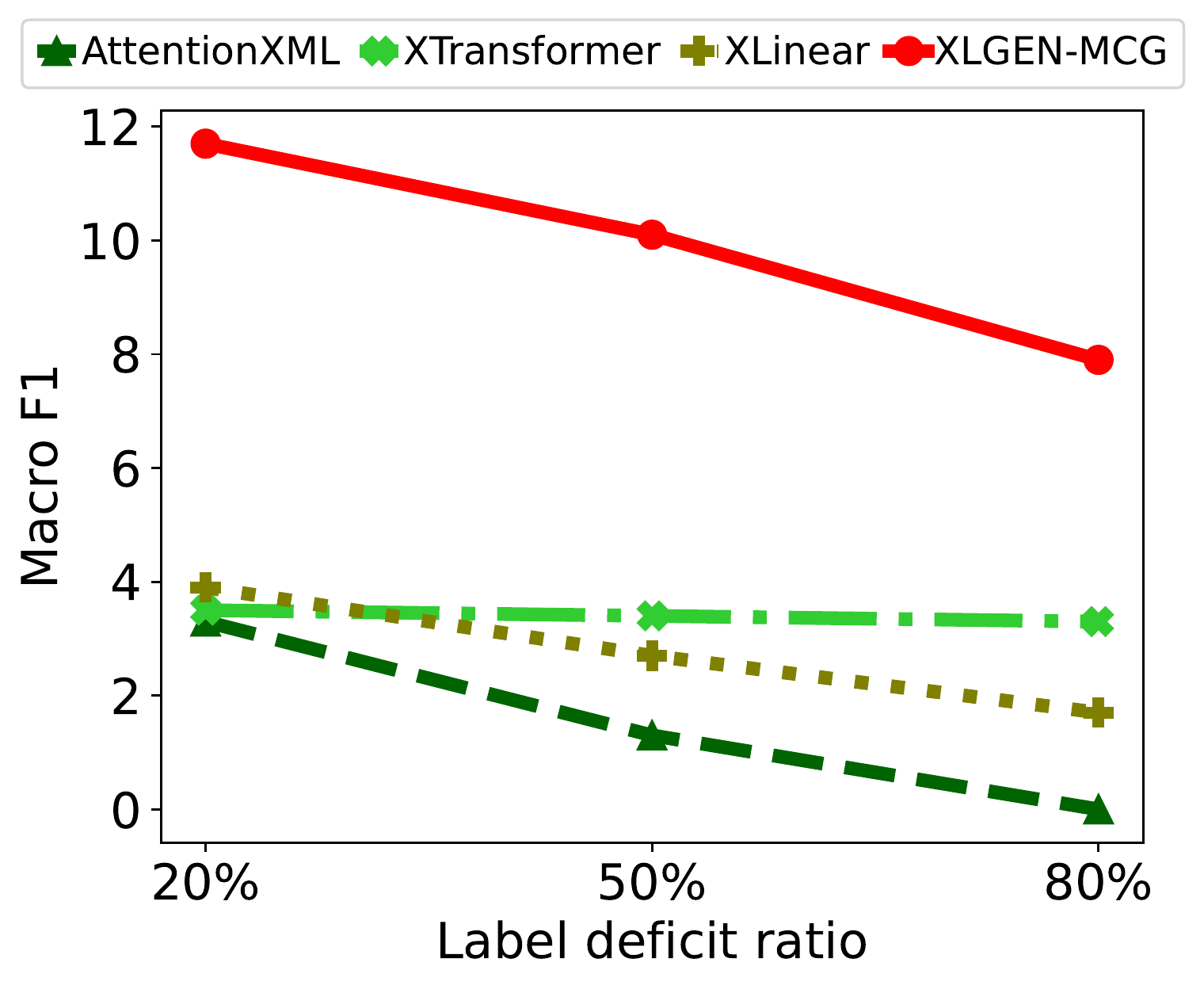}
}
\caption{\label{fig:chart_pu} 
Macro-averaged F1 scores in PU setting on \textsc{Wiki10-31K}.}
\end{figure}

\begin{figure*}[t!]
\center
{
\subfloat[Label Order]{\label{subfig:ablation_label}
\includegraphics[trim=0.3cm 0.2cm 0.7cm 0.4cm,clip,width=.32\linewidth]{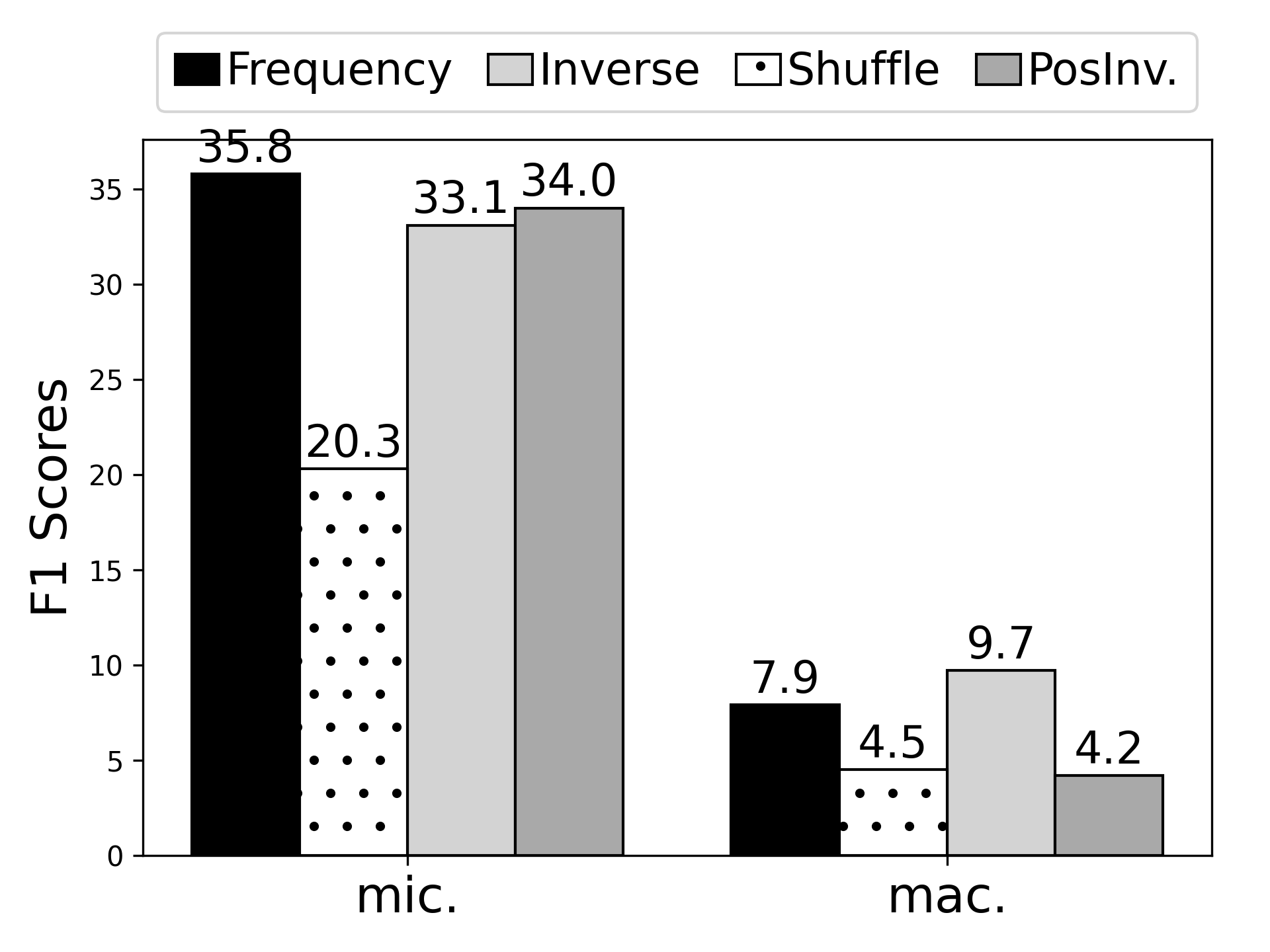}
}
\subfloat[Decoding Strategy]{\label{subfig:ablation_decoder}
\includegraphics[trim=0.3cm 0.2cm 0.7cm 0.4cm,clip,width=.32\linewidth]{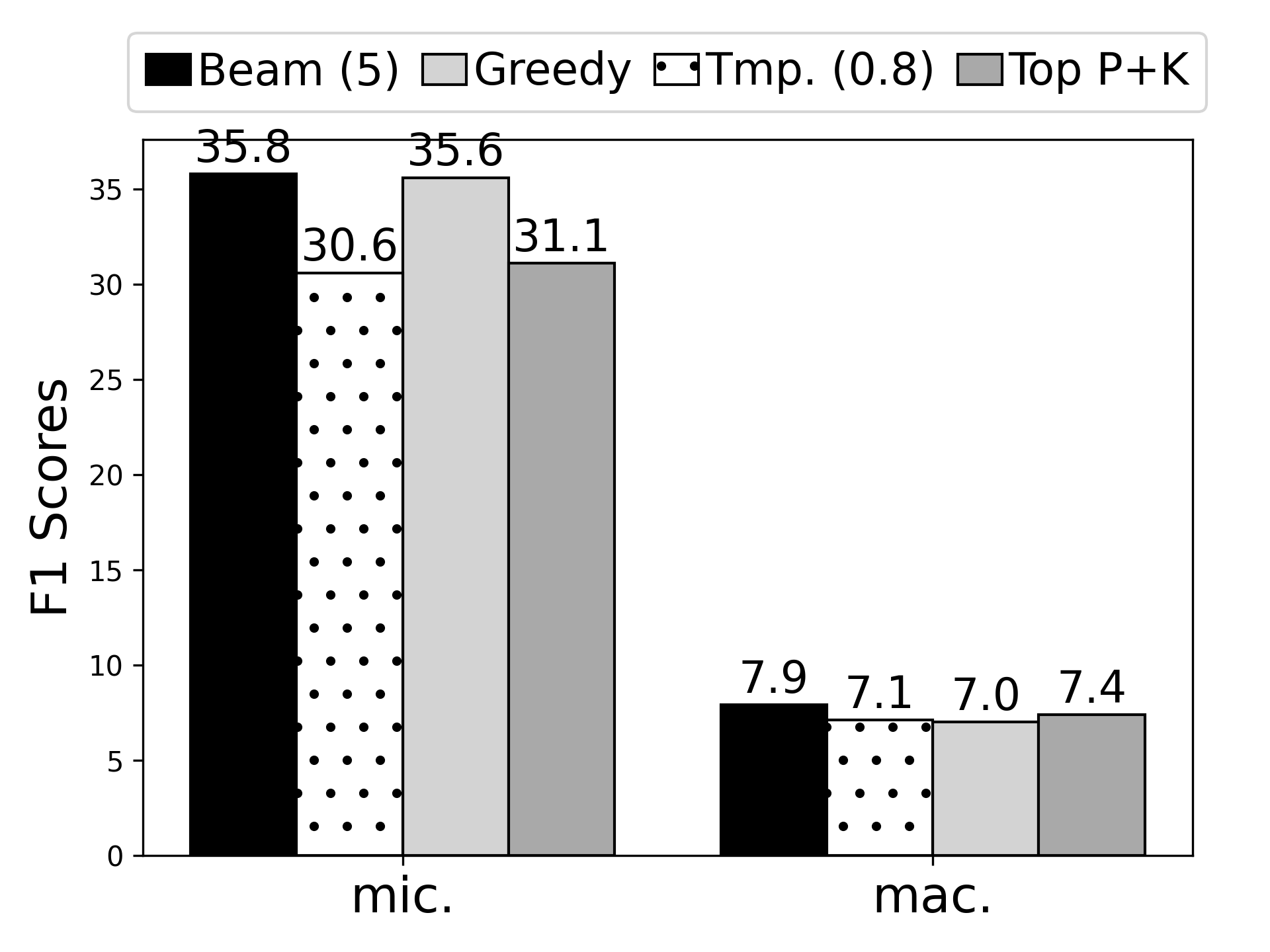}
}
\subfloat[Cluster Size]{\label{subfig:ablation_cluster}
\includegraphics[trim=0.3cm 0.2cm 0.6cm 0.4cm,clip,width=.33\linewidth]{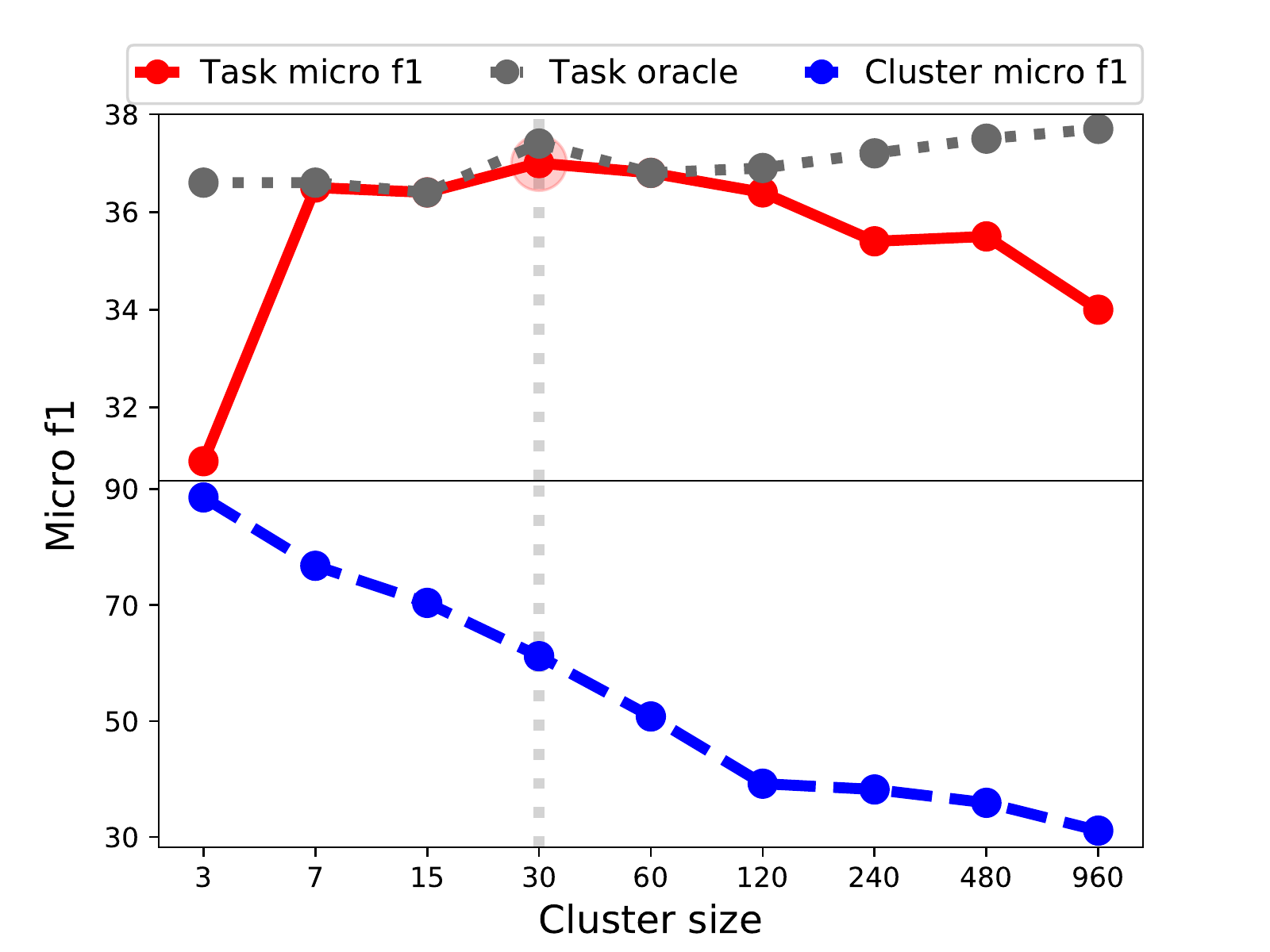}
}
}
\caption{\label{fig:ablation_study} Ablation study results on \textsc{Wiki10-31K}. (a) Performances of \method-\texttt{base} trained with various label orders. (b) Performances of \method-\texttt{base} trained with various decoding strategies.
(c) Cluster sizes vs \textcolor{red}{task} and \textcolor{blue}{clustering} performance.
We also report \textcolor{gray}{oracle} scores by using ground-truth cluster information in inference time.
%As cluster size increases, clustering performance decreases, while task performance is optimized with the cluster size around 30.
}
\end{figure*}

\paragraph{PU setting.}\label{subsec:pu}
In XMC, it is infeasible to annotate all relevant labels for an input text by checking every millions of labels. 
Therefore, many XMC datasets are indeed in PU setting. 
To evaluate the robustness against the positive and unlabeled properties, following \citet{hu2021predictive}, we make PU data setting by randomly eliminating positive labels for each instance with 50\% of deficit rate. 

As \method is trained with fewer positive labels in PU settings, the generated output labels tend to be fewer as well, causing lower recall than expected.
To increase the recall, we generate diverse label sequences using various sampling schemes in inference, which we call {\em ensemble generation}.
We combine generated results from three decoding strategies; beam search with size 5, Top $P+K$ sampling, and sampling with 0.8 temperature.

%In general, \method models outperform the baselines except for \textsc{EurLex-4K} with AttentionXML.
We find \method models outperform the baselines. Specifically, \method-\texttt{MCG} shows significantly strong scores, which indicates having predicted clusters as an additional input helps predict infrequent but correct labels.
In Figure~\ref{fig:chart_pu}, we additionally visualize macro F1 scores of PU settings on \textsc{Wiki10-31K} with various deficit rates.
Although \method-\texttt{MCG} drops with an increasing deficit rate, it still shows significant gaps with baseline scores.

%On the other hand, it is difficult to choose a single best setting among \method models since we utilize an {\em ensemble generation} that encourages randomnesses in label predictions.
%One observation is that \method-\texttt{MCG} shows strong macro F1 scores for all benchmarks except \textsc{AmznCat-13K}, which indicates having predicted clusters as an additional input helps predict infrequent but correct labels.

\begin{table}[t!]
\centering
\begin{adjustbox}{width=1.0\linewidth,center}
\begin{tabular}{@{}P{0.6cm}P{1.6cm}
P{0.9cm}P{0.9cm}P{0.9cm}P{0.9cm}}
\hline
&\multirow{2}{*}{}&\multicolumn{2}{c|}{\small{\textsc{EurLex-4K}}}
&\multicolumn{2}{c|}{\small{\textsc{Wiki10-31K}}}\\
\cline{3-6}
&&$Mic.$ & $Mac.$ 
&$Mic.$ & $Mac.$ \\
\hline
\multicolumn{2}{c}{\method-\texttt{BCL}}
& \textbf{60.7}
& \textbf{52.4}
& \textbf{37.7}
& \textbf{20.0}
\\
\hline
\multirow{3}{*}{GPT-3} & 
0-shot
& 9.2
& 6.3
& 7.6
& 4.5\\ 
& 1-shot
& 17.2
& 14.7
& 20.3
& 13.4
\\
& 5-shot
& 15.7
& 10.4
& 23.5
& 16.6 \\
\hline
\end{tabular}
\end{adjustbox}
\caption{Label performance of \method-\texttt{BCL} and in-context learning settings on 100 randomly selected samples.}
%For macro-averaged scores, we filter out labels that never occurred.}
\label{tab:performance_gpt3}
\end{table}

\subsection{Feasibility of in-context learning in XMC}\label{sec:feasibility_gpt3}
In-context learning \cite{brown2020language} shows a potential of generating unseen but positive labels as depicted in Figure \ref{fig:example_fincher}, such as ``geyser'' and ``physical\_reaction''.
% With recent advances in in-context learning, few-shot learning on top of pre-trained large language models is becoming increasingly popular across a variety of NLP domains.
In order to thoroughly validate the feasibility of in-context learning in XMC problems, we select 100 samples randomly from \textsc{EurLex-4K} and \textsc{Wiki10-31K}, and predict their labels using GPT-3~\citep{brown2020language} in zero/one/five-shot setups.
%using the OPenAI API~\footnote{\url{https://openai.com/api/}}.
We explore a few variations of prompts by tweaking label order or selecting few-shot examples differently, and report the best scores in Table~\ref{tab:performance_gpt3}.
%and find random sampling with decreasing frequency label order achieves the highest F1 scores.
%Table~\ref{tab:performance_gpt3} shows the performance comparison between prompt-based few-shot learning and \method. 
The performance of in-context learning significantly improves when we use more examples in the prompt, but they are far from the performance of \method.
Moreover, the performance gap between GPT-3 and \method is much larger in \textsc{EurLex-4K} where labels are formally annotated than in \textsc{Wiki10-31K} where labels are annotated by random users without a solid guideline. 
Unlike other multi-label classification tasks, XMC treats an extremely large number of labels, making it difficult to predict most unseen labels based on a few examples in in-context learning.
See \ref{appendix:supp_gpt3} for the details of the experimental setup for in-context learning and the performance comparison among prompt variations.

%See Appendix~ for the detail.\tj{Will add experimental details about prompt engineering and so on.}

\section{Ablation Study}
\label{sec:analysis}
We explore various factors that impact the performance of \method on \textsc{Wiki10-31K}, such as label orders (\S\ref{subsection:anal_label_ordering}) and sampling strategies (\S\ref{subsection:anal_decoding})
In order to reduce training costs, we mainly train \method-\texttt{base} on the base size model with epoch 5 for ablation tests.
We then investigate the model performance by clustering sizes and algorithms (\S\ref{subsec:anal_cluster}).

%%% Label Order performance table
% \begin{table}[t!]
% \center
% \small{
% \begin{tabular}{lcc|cc}
% % \multirow{2}{*}{\method-\texttt{base}}
% &\multicolumn{2}{c|}{\textsc{EurLex-4K}}
% &\multicolumn{2}{c}{\textsc{Wiki10-31K}}\\
% \cline{2-5}
% & \textit{Mic.} & \textit{Mac.}& \textit{Mic.} & \textit{Mac.}\\
% \hline
% Frequency & 58.0 &23.8&\textbf{35.8}&7.9\\
% Inverse &57.7&\textbf{24.7}&33.1&\textbf{9.7}\\
% Shuffle &56.3&22.0&20.3&4.5 \\ 
% \hline
% PosInv. &56.3&21.2&34.0&4.2 \\
% \hline
% \end{tabular}
% }
% \caption{Performances of \method-\texttt{base} trained with various label orders.
% }
% \label{tab:performance_order}
% \vspace{-2mm}
% \end{table}

\subsection{Label Orders}
\label{subsection:anal_label_ordering}

Label orders in decoder are important as \method sequentially generates labels. 
We compare three different label orders; label frequencies from high to low (Frequency), inverse label frequencies from low to high (Inverse), and shuffling where labels are randomly ordered per training epoch.
Inspired by~\citet{Lee2019SetTA}, we also consider ignoring label orders by resetting positional embeddings of each label as initial values in decoder~\footnote{But we keep the position embeddings for token sequences in one label to learn token positions.} which we call label positional invariant setting (PosInv.).
% We find trade-offs between macro and micro F1 scores between label frequencies and inverse label frequencies.
% For example, frequent label orders in general achieves the best micro F1 (Mi.) and ranking based F1 score (f@5) while inversely frequent label orders outperforms to the other label orders in terms of macro F1 (Ma.).
% It is because inversely frequent label orders make the model generate long-tail labels earlier with certainty, thus, per label scores in long-tail labels should be improved.

Figure~\ref{subfig:ablation_label} shows task performance across different label orders.
We find trade-offs between macro and micro F1 scores by the label frequency order (Frequency and Inverse) because inversely frequent label orders make the model generate long-tail labels earlier with certainty, thus, the scores of long-tail labels could improve.
On the other hand, shuffling~\citep{simig-etal-2022-open} crucially downgrades the performance since with randomly shuffled labels, \method tends to ignore co-occurrence patterns among labels in training time.
Also, we conjecture that positional invariant setting does not work well as it tweaks the original positional embeddings of pre-trained T5 model.

%% Decoding Strategy Performance Table
% \begin{table}[t!]
% \center
% \small{
% \begin{tabular}{lcc|cc}
% &\multicolumn{2}{c|}{\textsc{EurLex-4K}}
% &\multicolumn{2}{c}{\textsc{Wiki10-31K}}\\
% \cline{2-5}
% & \textit{Mic.}& \textit{Mac.} & \textit{Mic.} & \textit{Mac.} \\
% \hline
% Greedy & 57.5&23.3&35.6&7.0\\
% \hline
% Beam (3) & \textbf{58.0} &23.7&35.7&7.8 \\
% Beam (5) & \textbf{58.0} &\textbf{23.8}&\textbf{35.8}&\textbf{7.9}\\
% Beam (10) & 57.9 &\textbf{23.8}&35.4&7.8\\
% \hline
% Tmp. (0.8) &53.7&21.9&30.6&7.1 \\
% Top-$K$ (50) &51.7&20.8&28.7&6.8 \\ 
% Top-$P$ (0.9) &53.2&21.1&28.8&6.5 \\ 
% Top $P+K$ &53.6&21.5&31.1&7.4\\ 
% \hline
% \end{tabular}
% \caption{Performances of \method-\texttt{base} trained with different decoding strategies. 
% \vspace{-2mm}
% }
% \label{tab:performance_gen}
% }
% \end{table}

\subsection{Decoding Strategy}
\label{subsection:anal_decoding}
We now explore task performances with various sampling strategies in label generation.
We compare greedy search, beam search, sampling with restrictions such as Top-$K$~\citep{fan2018hierarchical} and Top-$P$~\citep{holtzman2019curious}, and sharpening vocabulary distributions with a temperature parameter.
% Table~\ref{tab:performance_gen} shows task performances across various decoding strategies with \method on \textsc{EURLex-4K} and \textsc{Wiki10-31K}. 
In Figure~\ref{subfig:ablation_decoder}, we find that beam search with size 5 achieves the best scores.
% Thus, we set up to generate labels with beam searching size 5 for the other experiments.
Interestingly, most sampling methods heavily degrade performances since our label spaces are not entirely open-ended.
% Compared to simply applying top-k or top-p samplings, integrating these two (e.g., Top-k+Top-p) or using temperatures while sampling improves label predictions.
We also explore ensemble methods to combine label outputs from different sampling strategies. 
Unlike the PU setting, however, they are not helpful in the full data setup since a sufficient number of labels are already generated by a single best generation strategy. 
Find the Appendix \ref{appendix:analysis_ensemble_generation} for details.

\subsection{Cluster Strategy}
\label{subsec:anal_cluster}
We show the effect of clustering algorithms and their parameters. 
We train \method-\texttt{MCG} fine-tuning T5-base with epoch 5.
We compare two clustering methods; K-means and Agglomerative clustering, and two text representations; TF-IDF and the recent T5 encoder.
We find K-means and pre-trained T5 encoder shows the best performance over other combinations, as described in Appendix~\ref{appendix:analysis_clustering_representation}.
% Thus, for all experiments with clustering information, we use K-means with pre-trained T5 encoder text representation.

% \dk{TODO: having too much references to Appendix might kill the paper. Later, merge appendix sections into one big section.}

% \textbf{Cluster Size.} 
Cluster size is another important factor for model performance. 
For example, a larger cluster size helps find label groups at a higher granularity, while it is much harder to be accurately predicted in inference time.
Here, we choose cluster size to be a power of two on average (e.g., around 30 containing 1024 labels for \textsc{Wiki10-31K} on average).
%~\footnote{Note that for main experiments in Table~\ref{tab:performance_total}, we find optimal cluster sizes via cross-validation.}.
% Here, we compare various cluster sizes from $\{\lceil \frac{L}{2^{k}} \rceil| k \in \{3,4,5,6,7,8,9,10\} \}$ so that each cluster has $2^{k}$ number of labels  on average.~\footnote{Note that for main experiments in Table~\ref{tab:performance_total}, we find optimal cluster sizes via cross-validation.} 
Figure~\ref{subfig:ablation_cluster} shows micro F1 scores of \method-\texttt{MCG} across cluster sizes in \textsc{Wiki10-31K}. 
% Two bar charts represent label~\ref{fig:chart_label} and cluster~\ref{fig:chart_cluster} macro-, micro- F1 scores.
Here we also report the upper bound of task performance (oracle) by using ground-truth cluster information.
As we expect, clustering performance decreases as the cluster size increases since it is much harder to predict clusters in a larger cluster space.
%\sj{predict what?}
%In particular, micro F1 scores for both datasets show a big drop from cluster size 120.
In terms of label prediction, we find that the model with smaller cluster sizes (e.g., $\leq$ 30) outperforms the larger ones, where the peak is around 30. 
% In particular, 
% Smaller cluster sizes (e.g., $\leq$30) model achieves better scores, where the peak is on 30 a. 
Although a larger cluster size helps elaborately specify labels in the same category, lower cluster prediction performance harms label performances as well and leads to a bigger performance gap compared with oracle task scores. 
%since wrongly predicted clusters are used for label inference.

\section{Qualitative Analysis}
\label{sec:qual_analysis}

Lastly, we evaluate the quality of generated labels via human evaluation (\S\ref{subsec:human}) and visualization of the semantic relations among labels (\S\ref{subsec:lab_semantic}).

\begin{table}[t!]
\centering
\small{
\begin{adjustbox}{width=\linewidth,center}
\begin{tabularx}{\linewidth}{
@{}cc 
P{0.5cm}P{0.5cm} 
@{\hskip 1.1cm}
P{0.5cm}P{0.5cm}@{}}
\hline
% \multicolumn{2}{c}{
% \multirow{2}{*}{\vtop{\hbox{\strut \textsc{Wiki10-31K} }}}} 
&
&\multicolumn{2}{@{}l}{AttentionXML}
&\multicolumn{2}{l}{\method-\texttt{BCL}}
\\
\cline{3-6}
&& \textbf{\#} & \textbf{\%}& \textbf{\#} & \textbf{\%} \\
\hline
\multirow{3}{*}{\makecell{Existing\\ labels}}&
\textit{Correct} &674&39.2\%&596&39.2\%\\
&\textit{Wrong} &776&45.1\%&457&30.0\%\\ 
&\textit{PU} &270&15.7\%&393&25.8\%\\ 
\hline
\multirow{2}{*}{\makecell{New \\labels}}&
\textit{Correct} &0&0.0\%&45&3.0\%\\
&\textit{Wrong} &0&0.0\%&30&2.0\%\\
\hline
\multicolumn{2}{c}{Total}  & 1,720 & 100\% & 1,521 & 100\%\\
\hline
\end{tabularx}
\end{adjustbox}
}
\caption{Human evaluation on \textsc{Wiki10-31K} having 1,720 true labels. The predicted labels are annotated and categorized to \textit{correct}, \textit{wrong}, and \textit{PU} labels, with their precision scores.
Note that \textit{correct} labels in the newly generated labels means they are possibly correct, according to the human annotators' decision.
}
\label{tab:qualitative_score}
\end{table}

\subsection{Human Evaluation}
\label{subsec:human}

The benchmark datasets have different label qualities.
For example, the labels from \textsc{EurLEX-4K}, annotated by the Publication Office of EU, are refined and structured while Wiki datasets are collaboratively labeled by general users, so the quality of labels is relatively lower than the other benchmarks.
Hence, we conduct human evaluations in both quantitative and qualitative ways to accurately measure the potential existence of PU labels and newly-generated labels.
In particular, we randomly select 100 instances from the test set and extract incorrectly predicted labels and/or newly predicted labels by \method and baseline models.
We then ask three human annotators to annotate and decide on possibly positive labels via majority voting.

Table~\ref{tab:qualitative_score} shows human evaluation results on the annotated \textsc{Wiki10-31K}.
Note that the number of predicted labels by \method-\texttt{BCL} is less than true labels because \method does not generate label with low confidence. 
In AttentionXML, on the other hand, we choose top-K labels as many as the number of true labels for each instance, so it has the same total labels as true labels.
% We first categorize predicted labels into three groups; correct labels existing in true set and incorrect / pu labels either existing in label spaces or newly generated.
% We then report precision scores for each of category.
Compared to the best baseline, AttentionXML, \method-\texttt{BCL} could generate more PU labels and reduce the number of wrong labels.
Also, our method generates 75 (=45+30) newly generated labels out of 1,521 where 60\% (=45/75) of them are correct, showing a relatively good generation quality of new labels.
Of course, we can control our model to only count the candidate labels and not any of these new labels for more accurate predictions, as measured in Table~\ref{tab:performance_full}.

\begin{table*}[t!]
\centering
\small
\begin{tabularx}{\textwidth}{@{}m{6.0cm}m{1.7cm}m{7.2cm}@{}}
\textbf{Input Document} & \textbf{Models} & \textbf{Labels} \\
\hline
\multirow{3}{\linewidth}{\setlength{\fboxsep}{0pt}\colorbox{yellow!90}{Emily Elizabeth Dickinson} (December 10, 1830– May 15, 1886) was an American poet.  Born in Amherst, Massachusetts to a successful family with strong community ties, she lived a mostly introverted and reclusive life. 
After she studied at the Amherst Academy for seven years in her youth, she spent a short time at ...} & True& 
authors biography dickinson emily journal library literature openaccess people poem poet poetry reference research to-read wiki wikipedia writers\\
\cdashline{2-3}
&AttentionXML & 
\textcolor{blue}{wiki poet writers wikipedia literature authors} 
\textcolor{red}{books} \sout{writing} \sout{history} \textcolor{red}{poets writer} \textcolor{blue}{people poetry biography} \sout{inspiration} \textcolor{blue}{american poems} \sout{luule} \\
\cdashline{2-3}
& \method-\texttt{BCL} &
\textcolor{blue}{wikipedia wiki people} \sout{art} \textcolor{red}{books} \textcolor{blue}{literature}
\sout{english}
\textcolor{blue}{poetry writers writer poet} \textcolor{red}{elizabeth} \textcolor{blue}{dickinson} \setlength{\fboxsep}{0pt}\colorbox{yellow!90}{emilydickinson}\\
\hline
\multirow{3}{\linewidth}{
Screenshot of vimeo.com home page \setlength{\fboxsep}{0pt}\colorbox{yellow!90}{Vimeo} is a video-centric social network site (owned by IAC/InterActiveCorp) which launched in November 2004. The site supports embedding, sharing, video storage, and allows user-commenting on each video page...} & True& articles computer reference socialnetworks technology tools video web2.0 wikipedia\\
\cdashline{2-3}
&AttentionXML & 
\textcolor{blue}{video web2.0 wikipedia}  \textcolor{red}{wiki} \sout{media} \sout{youtube} \textcolor{red}{videos} \sout{videoblogging} \sout{streaming}\\
\cdashline{2-3}
& \method-\texttt{BCL} &
\textcolor{blue}{wikipedia} \textcolor{red}{wiki} \textcolor{blue}{reference technology} \textcolor{red}{web internet} \sout{social} \textcolor{blue}{video web2.0} 
\sout{no\_tag} \textcolor{red}{socialnetworking} \sout{socialsoftware} \sout{phd} %\textcolor{red}{socialnet} 
\textcolor{red}{social\_networking social\_network} 
%\sout{ian\_minimalism} 
\setlength{\fboxsep}{0pt}\colorbox{yellow!90}{vimeo}
\\
\hline
\multirow{3}{\linewidth}{Diet Coke and Mentos Eruption is a reaction of Diet Coke and mint Mentos candies, a bottle of Diet Coke (other \setlength{\fboxsep}{0pt}\colorbox{yellow!90}{carbonated beverages} may be used instead) and dropping some Mentos. This causes the Coke to foam at a rapid rate and spew into the air... } & True&beverage candy chemistry coca-cola coke dietcoke drink eruption experiment experiments video explosion food fun funny interesting mint prank science \\
\cdashline{2-3}
&AttentionXML & 
\textcolor{red}{wikipedia} \textcolor{blue}{fun science} \sout{diet} \textcolor{red}{wiki} \textcolor{blue}{funny coke} \sout{tv} \textcolor{blue}{video} 
\sout{health} \textcolor{blue}{interesting humor food} \\
\cdashline{2-3}
& \method-\texttt{BCL} &
\textcolor{red}{wikipedia wiki} \textcolor{blue}{science interesting fun video funny food} \textcolor{red}{humor weird humour wtf} \sout{\#afterdarkclub} \setlength{\fboxsep}{0pt}\colorbox{yellow!90}{soda} \textcolor{blue}{eruption}\\
\hline
\end{tabularx}
\caption{Ground-truth and predicted labels from \method-\texttt{BCL} and AttentionXML on input documents in \textsc{Wiki10-31K}. 
% We ask human annotators to annotate positive (unlabelled) labels from incorrect and newly generated labels. 
We ask human annotators to annotate labels to be correct (\textcolor{blue}{blue}), wrong (\sout{strikethrough}), and PU (\textcolor{red}{red}). 
% For \method, among newly generated labels, we mark PU label and related context in input text with colorbox. (e.g., new PU label \setlength{\fboxsep}{0pt}\colorbox{yellow!90}{oscar} comes from the fact that David Fincher was academy-nominated.) 
For \method, we additionally mark potentially correct labels from the newly generated labels and their relevant contexts in input text with \setlength{\fboxsep}{0pt}\colorbox{yellow!90}{yellow} box. (e.g., a possibly correct label \setlength{\fboxsep}{0pt}\colorbox{yellow!90}{soda} is newly generated based on the fact that diet coke can be replaced with other carbonated beverage.) 
}
\label{tab:quant_anal}
\end{table*}

Lastly, we provide annotation examples in Table~\ref{tab:quant_anal}.
As we sort the label sequence by frequency in training for \method, frequently generated labels such as ``wikipedia" or ``wiki" are predicted first, followed by long-tail labels specified in the input text.
For AttentionXML, on the other hand, top predicted labels seem more aligned with the input context, although frequently generated labels still come in front.
Interestingly, new labels generated by \method come not only from the input context, but also previously generated labels.
% \sout{from the context of input text or}\jk{not only from the context of the input text but also} previously generated labels.
%For instance, on the wiki page of David Fincher, a new label ``oscar" is based on the input context that he is ``academy award-nominated"; here, \method even captures the fact that the word ``academy" can be a synonym of ``oscar".
For instance, on the Wikipedia page of diet coke and mentos eruption, a new label ``soda" is generated because input text contains ``carbonated beverages" which is synonym of ``soda".
On the Wiki page of Vimeo, on the other hand, after \method generates the PU label ``socialnetworking", followed by its synonyms such as ``social\_network" and ``social\_networking".

\subsection{Label Semantics in \method}
\label{subsec:lab_semantic}

\begin{figure*}
\centering
{
\subfloat[\label{subfig:emily} Emily Dickinson]{{\includegraphics[trim=0 0 0 0,clip,width=0.33\linewidth]{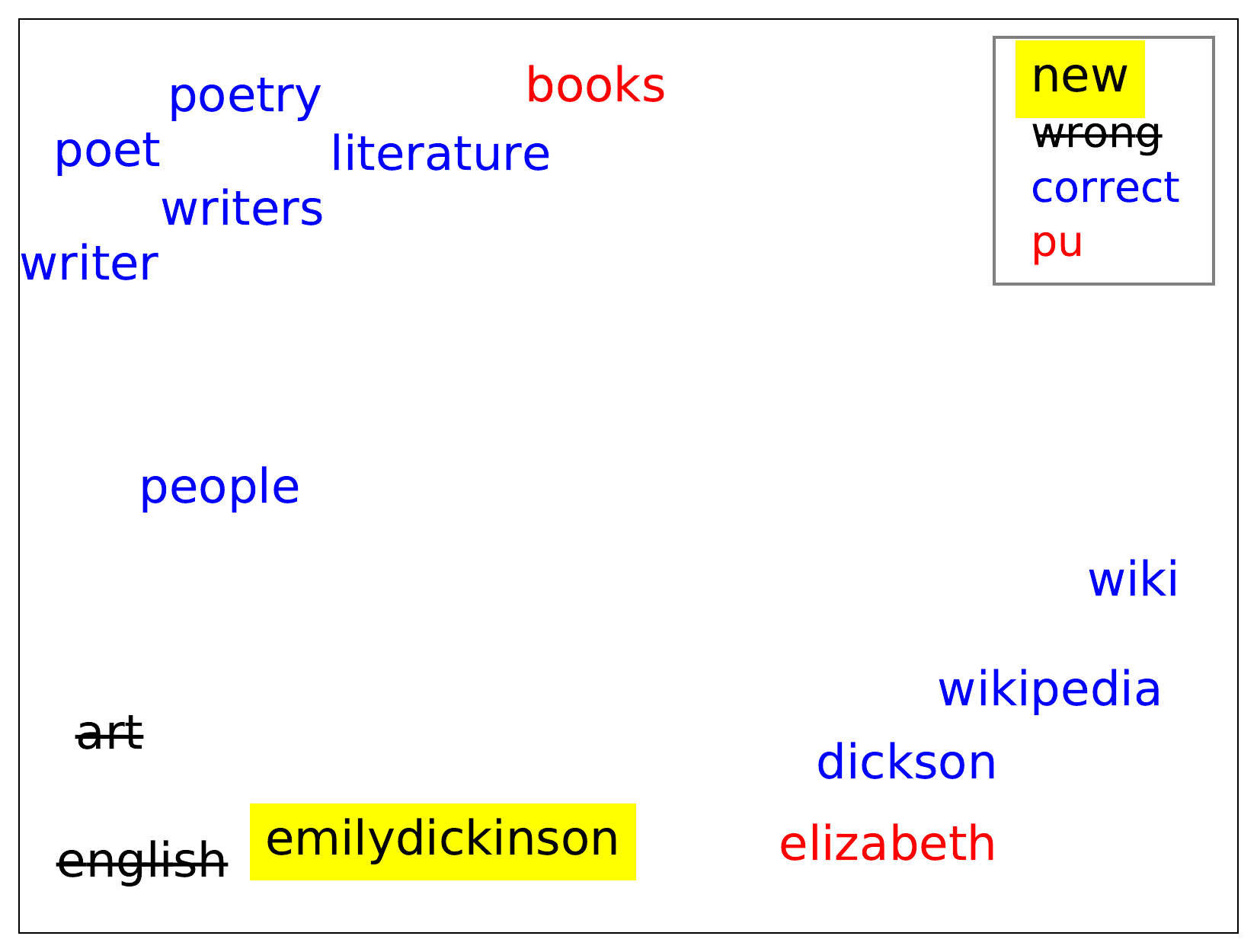}}
}
\subfloat[\label{subfig:vimeo} Vimeo.com]
{\includegraphics[trim=0 0 0 0,clip,width=0.33\linewidth]{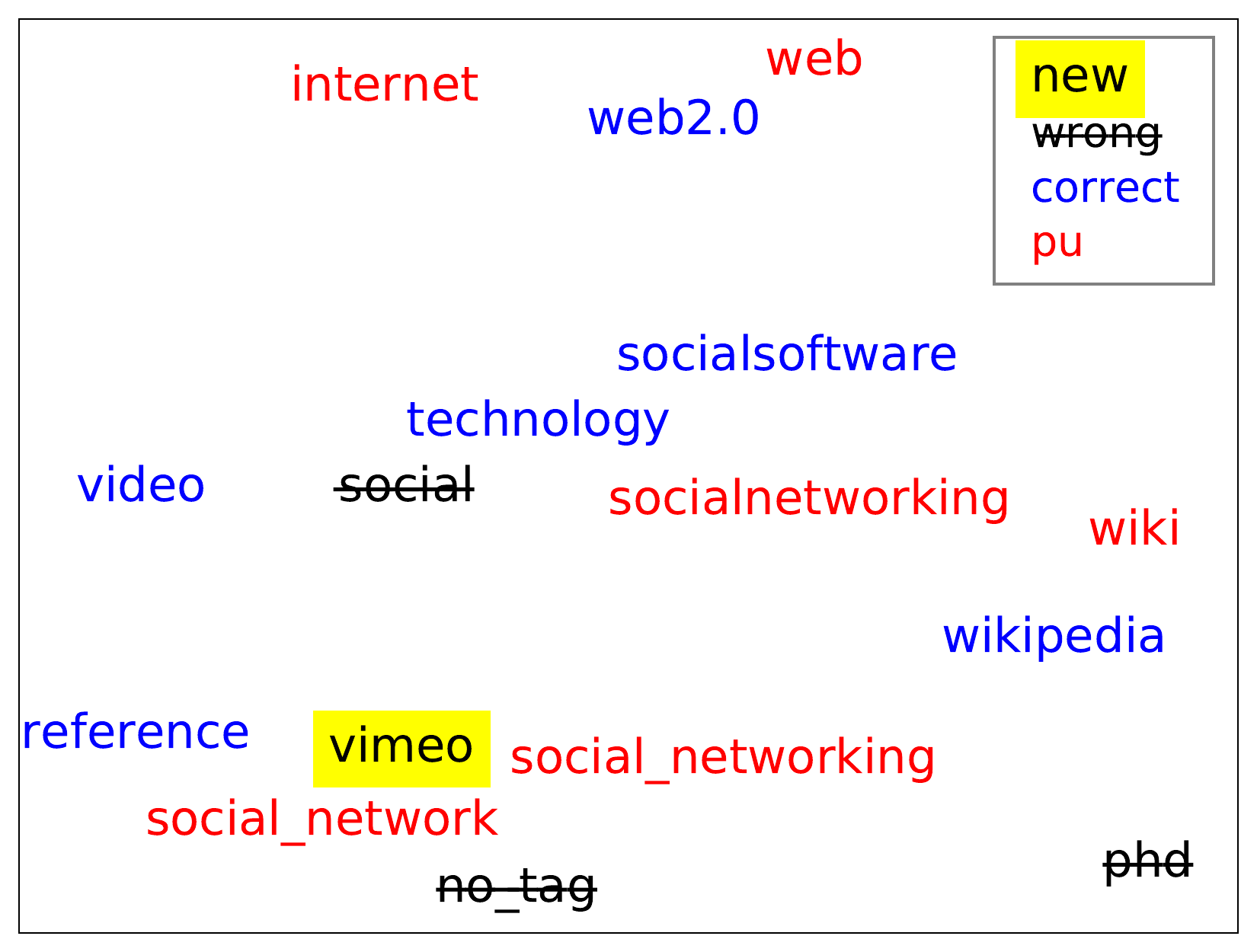}
}
\subfloat[\label{subfic:diet} Diet Coke and Mentos Eruptions]{{\includegraphics[trim=0 0 0 0,clip,width=0.33\linewidth]{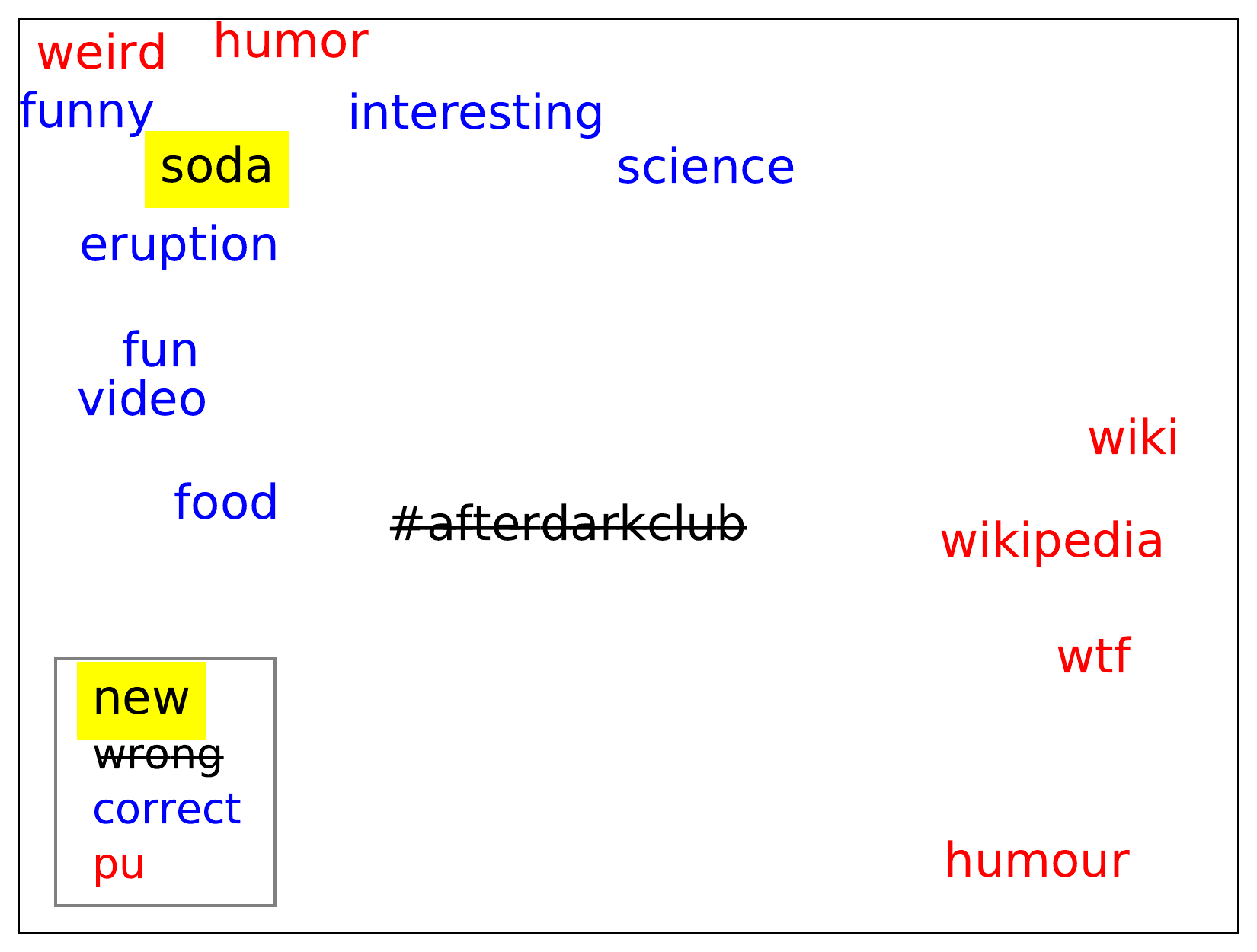}}
}
}
\caption{\label{fig:label_semantic} Visualization of generated labels by \method-\texttt{BCL} for the Wikipedia page examples in Table~\ref{tab:quant_anal}.
}
\end{figure*}

To better understand the semantics behind labels generated by \method,  we visualize an annotated labels of three examples from Table~\ref{tab:quant_anal} in Figure~\ref{fig:label_semantic}.
We get label embeddings from the last hidden state of the fine-tuned \method-\texttt{BCL} decoder and project them into two-dimensional T-SNE \cite{JMLR:v9:vandermaaten08a}.
If a single label is split by multiple tokens, we average the last hidden layers of all tokens.
We observe that frequently co-occurred labels (e.g., ``wiki"-``wikipedia" or ``weird"-``funny") have similar label embeddings.
Also, the newly generated labels become close to the co-occurred labels (e.g., ``soda" - ``funny" or ``eruption" in diet coke and mentos eruptions) via \method optimization.

\section{Conclusion} \label{sec:conclusion}
We apply text-to-text Transformers to extreme multi-label classification, by tweaking the classification problem as generation of label texts.
As we do not control the vocabulary space of generated labels, \method can create completely unseen but still relevant labels, inferred from the input context and semantic relationship from the previously generated labels.
Our experiments show that \method outperforms the classification baselines in general, and significantly improves the long-tail performance and PU setting.
Also, we observe utilizing label cluster information helps improve the performance in various settings.
\method is expected to more benefit from pre-trained models as they become larger and powerful \cite{kaplan2020scaling}.

\section*{Limitations and Future Directions}
First, we conduct our main experiments and additional analyses on certain languages such as English that has tremendous text corpora.
Extension to the low-resource languages might be challenging since this work requires text2text pre-trained models where those languages are applicable (e.g., multilingual T5 model), as well as the corresponding XMC datasets.

Also, compared to the efficient classification baselines, generative models are relatively expensive in terms of memory and time.
% For example, in \textsc{Eurlex-4K}, training a single epoch with T5-base model requires about 0.5 hrs in our setting.
For example, our experiment requires a lot of training resource as pre-trained models have >200 millions parameters to be tuned.
Thus, we use three p3.16xlarge AWS instances with 8 Nvidia V100 GPUs for training.
% For a practical use, applying an efficient version of transformers~\citep{tay2021scale} can help improve efficiency without big performance loss.
Using more efficient version of Transformers~\citep{tay2021scale} or applying distributed training should be considered for a resource reduction.

While in-context learning does not show comparable performance in XMC, we do observe that as the number of examples increases from zero to one to five, in-context learning can generate reasonable unseen but positive labels.
It would be interesting to explore the potential of in-context learning in XMC with more advanced prompting and example sampling in the future.

Lastly, the XMC task has a risk of being biased or overfit to small training datasets (e.g., \textsc{EurLex-4K} and \textsc{Wiki10-31K} contain only about 15,000 training examples). As with other commonly used NLP benchmarks, there is a potential risk that our proposed method may not work properly in the new test/train sets, though we anticipate that such a risk will be quite small. 

\section*{Ethics Statement}
We use the four XMC benchmark datasets which are publicly available and widely used in research~\footnote{\url{https://github.com/yourh/AttentionXML}}.
The datasets with social tags (e.g., \textsc{Wiki10-31K} and \textsc{Wiki-500k}) may contain inappropriate vulgarisms if they are not filtered out from the original data processing.
\bibliography{anthology,custom}
\bibliographystyle{acl_natbib}

\clearpage
\appendix 
\section{Appendix}
\label{sec:appendix}

% \subsection{Frequency Histograms of Labels}
% \label{appendix:label_hist}

% \begin{figure}[ht!]
% \centering
% {
% {\includegraphics[trim=0cm 0cm 0cm  0cm,clip,width=.95\linewidth]{img/label_freq.pdf}
% }
% }
% \caption{\label{fig:genXMC_freq} Frequency histograms of top-100 occurring labels in \textsc{EUR-Lex} (blue) and \textsc{Wiki10-31K} (red):
% The highly right-skewed histograms indicate the long-tailed labels in XMC.
% Only 0.47\% and 0.01\% of labels have more than 100 observations in \textsc{EUR-Lex} and \textsc{Wiki10-31K} train set, respectively.
% \vspace{-3mm}
% }
% \end{figure}
%\dk{Integrate two figures in 1 and log-scale... different colors... }

% Figure~\ref{fig:genXMC_freq} shows label frequency histograms for two XMC benchmarks.
% We observe the labels are heavily right skewed in the long-tail. 

\subsection{Details on Baseline Models}
\label{appendix:baseline}

AttentionXML~\citep{you2019attentionxml} is a label tree-based deep learning model. It uses a shallow and wide probabilistic label tree which allows to handle millions of labels and a multi-label attention mechanism by using raw text as input to capture the most relevant part of text to each label.

X-Transformer~\citep{chang2020taming} is the first scalable approach to apply deep transformer models in XMC task. 
In particular, it uses a pre-trained transformer encoder to assign labels to corresponding cluster. 
For each hierarchical cluster level, 
OVA classifiers are trained by only using sample instances under the same cluster, called teacher forcing negative (TFN) strategy. 
Unlike AttentionXML which only uses negative sampling, X-Transformer also uses the negative instances positively predicted by the classifier from the previous cluster level, called matcher-aware negatives (MAN).
Recently, \citet{zhang2021fast} proposed XR-Transformer to speed up X-Transformer's training time in recursive manner.
Thus, we use XR-Transformer instead of X-Transformer for the comparison.

XR-Linear~\citep{yu2020pecos} has a very similar architecture with XR-Transformer, except that it only uses simple tf-idf text features instead of transformer encoder outputs. 
For OVA classification, linear matchers recursively solve XMC sub-problem for each hierarchical cluster level.

In order to fit score outputs into [0,1], we apply sigmoid post processor implemented by the authors for XR-Transformer and XR-Linear.

\subsection{Details on \method Training}
\label{appendix:hyperparam}
\begin{table}[ht!]
\begin{adjustbox}{width=0.9\linewidth,center}
\begin{tabular}{l|ccc}
\hline
& \method-\texttt{BCL} & %\method-\texttt{BCL}$_{concat}$ &
\method-\texttt{MCG}\\
\hline
\textsc{EurLex-4K} & 80 &20 \\ %&40
\textsc{AmznCat-13K} &80 &20\\ % &80
\textsc{Wiki10-31K} &60 &20  \\ %&40 
\textsc{Wiki-500K} &80&20 \\ %&40
\hline
\end{tabular}
\end{adjustbox}
\caption{Optimal cluster sizes for the \method training.}
\label{tab:cluster_size}
\end{table}

We finetune the T5-large (\textsc{EurLex-4K, Wiki10-31K}) or the T5-base (\textsc{AmazonCat-13K, Wiki-500K}), with epoch 10 (\textsc{EurLex-4K, AmazonCat-13K}) or epoch 5 (\textsc{Wiki10-31K, Wiki-500K}) based on the data and/or label size.

We set up input length as 500 for all benchmark datasets and use different output length based on the label lengths in train set; 90 for \textsc{EurLex-4K} and 165 for other three benchmarks.
We optimize \method using AdamW~\citep{Loshchilov2019DecoupledWD} with learning rate 2e-4. 
%\tj{@JK, could you add hyper-parameters below in this section? Also, please add whatever else if exists!}
%\begin{itemize}
%    \item input length
%    \item output length
%    \item batch size
%\end{itemize}

For \method-\texttt{BCL}, we set up an initial weight value $\lambda$ as 1.0 and reduce it to $\frac{1}{k}$ for every epoch number $k$.

For cluster-based \method architectures, we train k-means clustering and optimize the cluster size via cross-validation from the range of \{10,20,30,...,100\}.
In Table~\ref{tab:cluster_size}, we report optimal cluster sizes for \method training.

Note that each of T5 models have 220 million (T5-base) or 770 million (T5-large) parameters to be tuned.
Also for training, we use a small batch size (1) since pre-trained T5 models are large to be fitted in a single GPU machine. 
Due to the model size, we use two GPU machines via model parallelism for T5-large and a single GPU machine for T5-base in training.
%In particular, for T5-base, we use a single GPU machine and for T5-large, we use two GPU machines via model parallelism.
Also, due to the training cost and time, we report the performance scores from the single running of training and inference.
We basically modify the T5 code from huggingface library~\footnote{\url{https://huggingface.co/}}, and our code will be publicly available at \url{https://github.com/alexa/xlgen-eacl-2023}.

% \textbf{\method inference.}
% For inference, we use a single generation strategy with beam search with size 5.
%For \method-\texttt{BCL}$_{concat}$, we use a predicted clusters in inference time. By default, we set up threshold for clusters as 0.5.
% For PU settings which contain much fewer positive labels than the original ones, the cluster confidences are biased to be lower. %Therefore, we set lower thresholds, 0.3 and 0.4 for PU deficit ratio 20\% and 50\%, respectively.

\begin{table}[t!]
\centering
\begin{adjustbox}{width=1.0\linewidth,center}
\begin{tabular}{@{}P{0.8cm}P{1.8cm}P{1.2cm}
P{0.9cm}P{0.9cm}P{0.9cm}P{0.9cm}}
\hline
&\multirow{2}{*}{Sample}&\multirow{2}{*}{Label}&\multicolumn{2}{c|}{\small{\textsc{EurLex-4K}}}
&\multicolumn{2}{c|}{\small{\textsc{Wiki10-31K}}}\\
\cline{4-7}
&&&$Mic.$ & $Mac.$ 
&$Mic.$ & $Mac.$ \\
\hline
\multirow{4}{*}{0-shot} & 
Random & Random & 5.3 & 3.8 & 7.1 & 3.8\\ 
& Random & Frequency & \textbf{9.2} & \textbf{6.3} & \textbf{7.6} & \textbf{4.5}\\ 
& Most Label & Random & 6.0 & 4.3 & 7.2 & 4.4\\ 
& Most Label & Frequency & 5.0 & 3.5 & 6.7 & 3.5\\
\hline
\multirow{4}{*}{1-shot} & 
Random & Random & 14.8 & 10.7 & 13.6 & 11.8 \\
& Random & Frequency & 16.1 & 10.4 & \textbf{20.3} & 13.4\\ 
& Most Label & Random & \textbf{17.2} & \textbf{14.7} & 17.9 & \textbf{16.8}\\ 
& Most Label & Frequency & 15.1 & 12.2 & 18.1 & 16.1\\
\hline
\multirow{4}{*}{5-shot} & 
Random & Random & \textbf{15.7} & 10.4 & 17.9 & 14.2 \\
& Random & Frequency & 13.1 & 9.7 & \textbf{23.5} & 16.6 \\
& Most Label & Random & 11.0 & 9.6 & 19.8 & \textbf{18.0}\\ 
& Most Label & Frequency & 12.5 & \textbf{11.3} & 21.5 & 15.1\\
\hline
\end{tabular}
\end{adjustbox}
\caption{Micro-averaged and macro-averaged F1 scores in-context learning settings on 100 randomly selected samples.
We test two label ordering strategies, random and decreasing label frequency (frequency), as well as two sampling strategies, random and selecting examples with the most labels (most label).
The highest scores are \textbf{bold}.}
%For macro-averaged scores, we filter out labels that never occurred.}
\label{tab:performance_gpt3_app}
\end{table}

\begin{table*}[ht!]
\begin{adjustbox}{width=0.99\linewidth,center}
\begin{tabular}{lc|c|c|c}
\hline
\multirow{3}{*}{}&\textsc{EurLex-4K}&\textsc{AmznCat-13K}&{\textsc{Wiki10-31K}}&{\textsc{Wiki-500K}}\\
\cline{2-5}
& F@1/F@3/F@5/F@10
& F@1/F@3/F@5/F@10
& F@1/F@3/F@5/F@10
& F@1/F@3/F@5/F@10\\
\hline
XR-Transformer 
& \textbf{27.4}/47.5/47.2/34.7 
& \textbf{31.6}/55.2/53.5/38.0 
& \textbf{8.8}/20.3/24.4/23.8 
& 24.1/34.2/32.8/25.5\\
XR-Linear 
& 26.1/47.8/51.1/41.2 
& 30.5/55.8/58.6/45.7
& 8.5/19.2/24.9/29.3 
& 22.8/32.5/31.3/24.4 \\ 
AttentionXML 
& 26.9/\textbf{51.9}/\textbf{58.8}/59.9 
& 30.8/\textbf{59.2}/67.0/ 69.9
& 8.6/\textbf{21.0}/\textbf{28.1}/\textbf{35.6}
& \textbf{24.4}/42.9/48.9/52.8\\
\hline
\method-\texttt{base}
& 21.3/46.9/57.8/59.8 
& 30.8/57.4/65.3/69.0 
&  8.1/15.5/21.7/30.1
& 23.7/\textbf{44.0}/50.4/54.6 \\ 
\method-\texttt{BCL} 
& 21.2/47.2/58.7/\textbf{60.7} 
& 31.0/57.7/65.5/69.2 
& 8.1/15.4/21.5/30.1 
& 23.8/43.8/50.3/54.6 \\
% \method-\texttt{BCL}$_{concat}$  
% & 21.4/47.4/58.5/60.5 
% & 31.1/56.0/63.0/65.8
% & 8.1/16.0/22.5/31.5
% & 23.5/43.3/49.7/53.9\\
\method-\texttt{MCG} 
& 20.9/47.0/58.1/60.2 
& 31.2/58.8/\textbf{67.5}/\textbf{71.2}
& 8.1/15.6/22.2/31.2
& 23.8/\textbf{44.0}/\textbf{50.5}/\textbf{54.8} \\
\hline
\end{tabular}
\end{adjustbox}
\caption{Supplementary scores on benchmark datasets. 
We report ranking-based @k (k=1,3,5,10) F1 scores (F@k) as supplementary metrics.
The highest scores are \textbf{bold}.}
\label{tab:performance_supp}
\end{table*}

\begin{table*}[ht!]
\begin{adjustbox}{width=0.99\linewidth,center}
\begin{tabular}{lccc|ccc|ccc}
\hline
\multirow{3}{*}{}&\multicolumn{3}{c|}{\textsc{EurLex-4K}}
&\multicolumn{3}{c|}{\textsc{Wiki10-31K}}
%&\multicolumn{3}{c|}{\textsc{AmznCat-13K}}
&\multicolumn{3}{c}{\textsc{Wiki-500K}}\\
\cline{2-10}
&{0-shot} &{1-shot}&{5-shot} 
&{0-shot} &{1-shot}&{5-shot} 
%&{0-shot} &{1-shot}&{5-shot} 
&{0-shot} &{1-shot}&{5-shot} \\
\hline
XR-Transformer & 0.0/0.0 & 1.5/0.5 & 4.7/2.3& 0.0/0.0 & 2.5/1.6 & 2.9/1.7 
 %& -/- &  0.0/0.0 & 0.2/0.1 
 & 0.0/0.0 & 0.0/0.0 &  0.0/0.0 \\
XR-Linear &  0.0/0.0 & 5.9/1.1 & 8.6/2.7 & 0.0/0.0 & 2.8/2.3 & 2.9/2.4  
%&  -/- &  0.0/0.0 & 2.1/1.1 
& 0.0/0.0 & 0.2/0.1 & 1.4/0.9\\ 
AttentionXML & 0.0/0.0 & 16.3/2.4 & 28.4/8.3 & 0.0/0.0 & 0.3/0.2 & 5.6/1.9 
%& -/- &  0.0/0.0 &  1.9/0.9 
& 0.0/0.0 & 2.2/1.3 & 16.1/9.2\\
\hline
\method  & 5.3/3.2 & 21.4/3.5 & 34.9/10.8 &4.5/2.9 & 14.4/\textbf{8.4} & \textbf{17.9}/\textbf{7.5}
%& -/- & \textbf{15.4}/\textbf{9.0} & \textbf{27.1}/\textbf{17.1}
&\textbf{21.6}/22.5 & 35.6/24.1 & 39.6/28.7\\ 
\method-\texttt{BCL} & \textbf{7.3}/4.3 & \textbf{25.0}/\textbf{4.1} & \textbf{36.1}/\textbf{11.4} &\textbf{5.0}/3.3 & \textbf{14.5}/\textbf{8.4} & 17.6/7.3  
%& -/- & 13.4/7.9 & 25.8/15.9 
&18.6/23.2 & 36.2/24.8 & 40.2/29.5 \\
% \method-\texttt{BCL}$_{concat}$  & 6.3/3.7 & 21.3/3.6 & 35.9/10.9 &\textbf{6.2}/4.1 & \textbf{15.3}/\textbf{9.0} & \textbf{18.8}/\textbf{7.8}  
%  & -/-  & 11.8/6.7 & 23.5/14.4 &17.4/20.0 & 33.3/22.4 & 37.8/27.3  \\
\method-\texttt{MCG} & \textbf{7.3}/\textbf{4.5} & 16.7/2.7 & 35.9/11.1 &\textbf{5.0}/\textbf{11.1} & 13.9/8.1 & 17.3/7.2   
%& -/- &  12.2/6.5 & 25.0/14.6 
&20.7/\textbf{23.7} & \textbf{37.0}/\textbf{25.5} & \textbf{40.6}/\textbf{29.9}  \\ 
\hline
\end{tabular}
\end{adjustbox}
\caption{Task performances on benchmark datasets in full few-shot setup. 
We use conventional micro-averaged ($Mic.$) and macro-averaged ($Mac.$) F1 scores and mark $Mic.$/$Mac.$ in the table.
The highest scores are \textbf{bold}.}
\label{tab:performance_fewshot_app}
\end{table*}

\begin{table*}[ht!]
\centering
\begin{adjustbox}{width=0.99\linewidth,center}
\begin{tabular}{lccc|ccc|ccc}
\hline
\multirow{3}{*}{}&\multicolumn{3}{c|}{\textsc{EurLex-4K}}
%&\multicolumn{3}{c|}{\textsc{AmznCat-13K}}
&\multicolumn{3}{c|}{\textsc{Wiki10-31K}}
&\multicolumn{3}{c}{\textsc{Wiki-500K}}\\
\cline{2-10}
&\multicolumn{3}{c|}{Positive Unlabeled Deficit Ratio}
%&\multicolumn{3}{c|}{Positive Unlabeled Deficit Ratio}
&\multicolumn{3}{c|}{Positive Unlabeled Deficit Ratio}
&\multicolumn{3}{c}{Positive Unlabeled Deficit Ratio}\\
\cline{2-10}
&{$20\%$}&{$50\%$}&{$80\%$}
%&{$20\%$}&{$50\%$}&{$80\%$}
&{$20\%$}&{$50\%$}&{$80\%$} &{$20\%$}&{$50\%$}&{$80\%$} \\
\hline
XR-Transformer
& 32.6/10.9 &  25.9/8.6 & 14.6/4.7 
%& 45.7/13.8 & 32.1/13.3 & 29.2/12.5 
& 20.7/3.5& 16.7/3.4&12.0/3.3
& 29.2/8.0 & 24.5/7.0 & 17.7/5.1 \\
XR-Linear
& 38.6/12.1 & 27.8/8.8 & 12.5/3.7
%& 47.2/18.3 & 36.1/11.8 & 18.5/6.8
& 16.1/3.9& 11.5/2.7 & 5.1/1.7
& 14.4/3.7 & 9.9/2.8 & 5.9/2.0\\ 
AttentionXML 
& \textbf{57.6}/23.6 & \textbf{52.3}/18.7 & \textbf{40.0}/11.1
%& 65.6/18.7 & 61.8/14.6 & \textbf{56.7}/8.8
& \textbf{34.7}/3.3 & 27.7/1.3 & 14.1/0.0
& 50.9/18.9 & 46.0/11.3 & 35.5/5.0 \\
\hline
\method-\texttt{base} 
& 55.7/24.4 & 47.5/18.8 & 31.3/9.8
%& \textbf{70.0}/\textbf{49.1} & \textbf{67.8}/\textbf{41.8} & 53.7/\textbf{27.6}
& 33.6/9.1 & 32.3/7.7 & 22.7/2.8
& \textbf{51.1}/\textbf{37.3} & \textbf{48.7}/31.6 & \textbf{37.9}/\textbf{25.4}\\ 
\method-\texttt{BCL}
& 56.0/24.4 & 47.9/19.3 & 31.4/10.0
%& 69.9/48.2  & 67.7/41.1 & 53.6/27.0
& 33.2/9.1 & \textbf{33.0}/8.0 & 23.6/2.9
& 50.8/37.0 & 48.5/31.4 & 37.8/25.2 \\
%\method-\texttt{BCL}$_{concat}$
% & 55.8/23.9 & 47.7/18.8 & 32.8/10.2
% & 68.9/45.6 & 65.8/39.1 & 54.2/\textbf{27.9}
% & 34.0/9.1 & 32.6/7.7 & \textbf{25.1}/3.2
% & 50.8/36.5 & 48.2/31.2 & \textbf{39.4}/\textbf{26.4}\\
\method-\texttt{MCG}
& 55.5/\textbf{27.8} & 48.2/\textbf{21.2} & 32.6/\textbf{13.3}
%& 68.9/45.8 & 66.1/38.3 & 52.2/23.4
& 32.4/\textbf{11.7} & \textbf{33.0}/ \textbf{10.1} & \textbf{24.0}/\textbf{7.9}
& 50.7/\textbf{37.3} & 48.5/\textbf{32.7} & 35.9/24.8 \\ 
\hline
\end{tabular}
\end{adjustbox}
\caption{Task performances on benchmark datasets in full PU setup. 
We use conventional micro-averaged ($Mic.$) and macro-averaged ($Mac.$) F1 scores and mark $Mic.$/$Mac.$ in the table.
The highest scores are \textbf{bold}.}
\label{tab:performance_total_app}
\end{table*}

\subsection{In-context learning in XMC}
\label{appendix:supp_gpt3}

\begin{figure*}[t!]
%\hspace*{-0.5cm}
\centering
\subfloat[\label{fig:gpt_rand_rand} Label frequency order with random sample]{\includegraphics[trim=0cm 1cm 0cm  7cm,clip,width=.5\linewidth]{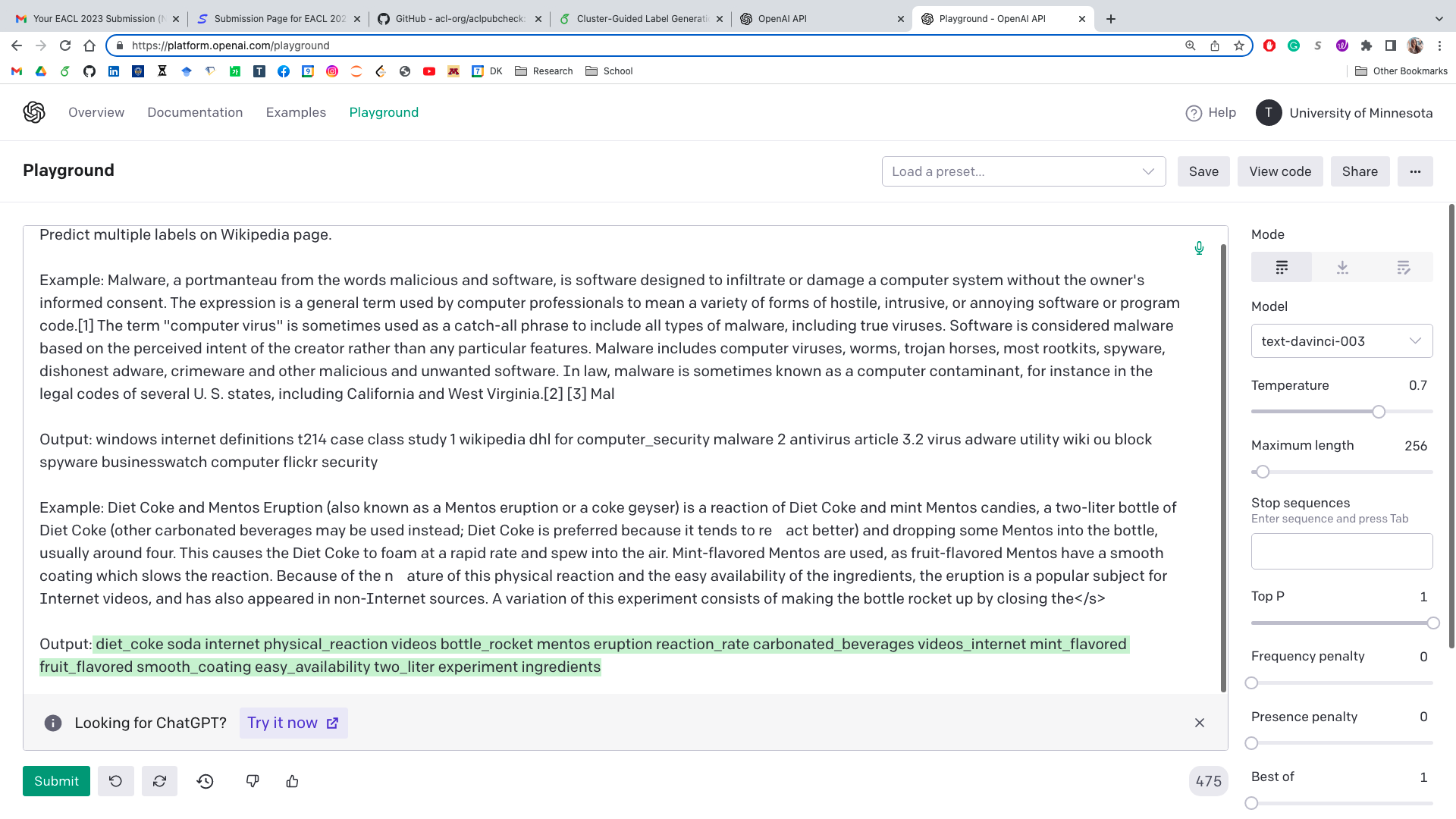}}
\subfloat[\label{fig:gpt_freq_rand} Label random order with random sample]{\includegraphics[trim=0cm 1cm 0cm  7cm,clip,width=.5\linewidth]{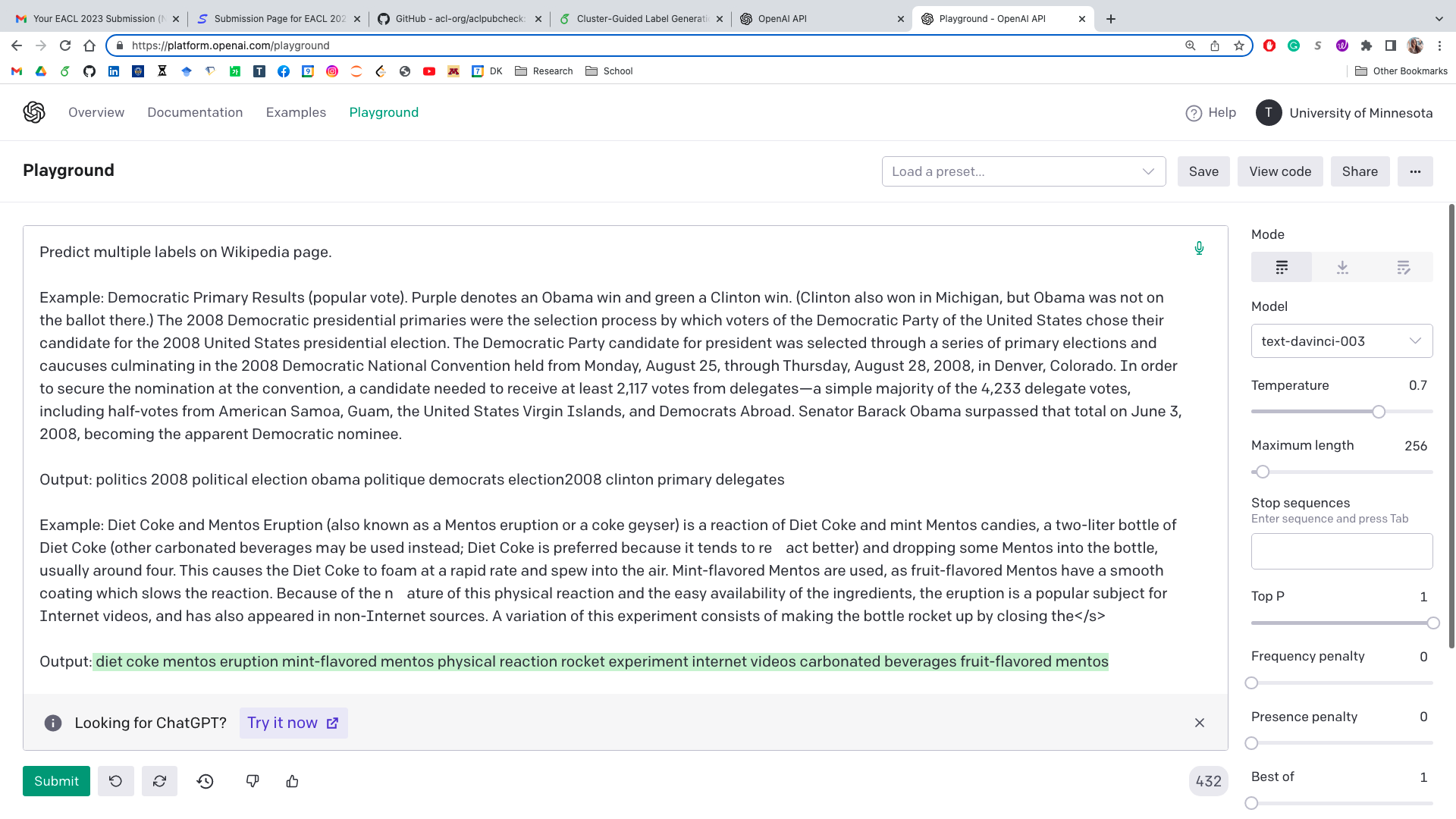}}
\caption{\label{fig:gpt_prompt} Input prompt and generated outputs (green-shaded) for Wikipedia page of Elizabeth Dickinson with 1-shot example.}
\end{figure*}

For in-context learning, we use OpenAI GPT-3 text-davinci-002 model with temperature 0.7 and max tokens 256.
To find the optimal prompt, we use prompt variations with different label orders and few-shot example sampling strategy.
Furthermore, we test two label ordering strategies, random and decreasing label frequency, as well as two sampling strategies, random and selecting examples with the most labels.
See Figure~\ref{fig:gpt_prompt} for GPT-3 prompt input and generated output.
Table~\ref{tab:performance_gpt3_app} shows the in-context learning performances across different label ordering and sampling strategies.
The best macro-averaged F1 scores for \textsc{Wiki10-31K} are achieved with label frequency ordering with random sampling; however, there is no consistently outperforming strategy for \textsc{EurLex-4K}.

\subsection{Additional Task Performances on Benchmark Datasets}
\label{appendix:supp_score}
For the full setup, we also report ranking based scores in Table~\ref{tab:performance_supp}.
In general, for supplementary metrics (F@k) \method shows comparable results with baselines except F@1, and F@3 in \textsc{EurLex-4K} and \textsc{Wiki10-31K}.
Note that for \method, we just treat the order of generated labels as a rank, which might \textbf{not} be correct since such generated labels should have a equal priority in theory.
For this reason, \method has lower F@k scores with smaller k.
However, such score gaps between baselines and \method decrease as k increases, like \textsc{EurLex-4K} with \method-\texttt{BCL}, or even \method achieves higher performances in the larger benchmarks (e.g., F@5 and F@10 in \textsc{AmznCat-13K} and \textsc{Wiki-500K}).
Also, full micro/macro F1 scores for tail labels and PU settings are in Table~\ref{tab:performance_fewshot_app} and Table~\ref{tab:performance_total_app}, respectively.
% and PU setups are in Table~\ref{tab:performance_fewshot_app} and Table~\ref{tab:performance_total_app}, respectively.

\subsection{Additional Analyses on Base Model Comparison}
\label{appendix:analysis_base_comparison}

\begin{table}[ht!]
\center
\small{
\begin{tabular}{lcc|cc}
\hline
\multirow{2}{*}{\method-\texttt{base}}&\multicolumn{2}{c|}{\textsc{EurLex-4K}}
&\multicolumn{2}{c}{\textsc{Wiki10-31K}}\\
\cline{2-5}
& \textbf{$Mic.$} & \textbf{$Mac.$}
& \textbf{$Mic.$} & \textbf{$Mac.$}\\
\hline
T5-base & \textbf{58.0}&\textbf{23.8}
&\textbf{35.8}&\textbf{7.9}\\
BART-base &55.6&23.4&35.0&7.7 \\ 
\hline
\end{tabular}
}
\caption{Task performances of \method-\texttt{base} trained with different pre-trained model architectures on \textsc{EurLex-4K} and \textsc{Wiki10-31K}.
%We report macro / micro averaged F1 scores ($Mic.$ / $Mac.$) and ranking based F1 score with k =5.
The highest scores are \textbf{bold}.
\vspace{-3mm}}
\label{tab:performance_bart}
\end{table}

For \method, we can use any pre-trained text-to-text models. 
We compare task performance of two popular text-to-text models in Table~\ref{tab:performance_bart}; T5~\citep{raffel2019exploring} and BART~\citep{lewis2020bart} by finetuning \method-\texttt{base}.
In general T5 model outperforms BART, therefore, we use pre-trained T5 architectures for our main experiments.

\begin{figure*}[ht!]
\centering
\subfloat[\label{fig:ex_1} Emily Dickinson]
{\includegraphics[trim=0 0 0 0,clip,width=0.33\linewidth]{img/tsne_Wiki10-31K_bce-sep_kmeans_t5-large_40_2.pdf}
}
\subfloat[\label{fig:ex_2} Vimeo.com]
{\includegraphics[trim=0 0 0 0,clip,width=0.33\linewidth]{img/tsne_Wiki10-31K_bce-sep_kmeans_t5-large_40_25.pdf}
}
\subfloat[\label{fig:ex_3} Diet Coke and Mentos Eruption]
{\includegraphics[trim=0 0 0 0,clip,width=0.33\linewidth]{img/tsne_Wiki10-31K_bce-sep_kmeans_t5-large_40_102.pdf}
}
\\
\subfloat[\label{fig:ex_4} David Fincher]
{\includegraphics[trim=0 0 0 0,clip,width=0.33\linewidth]{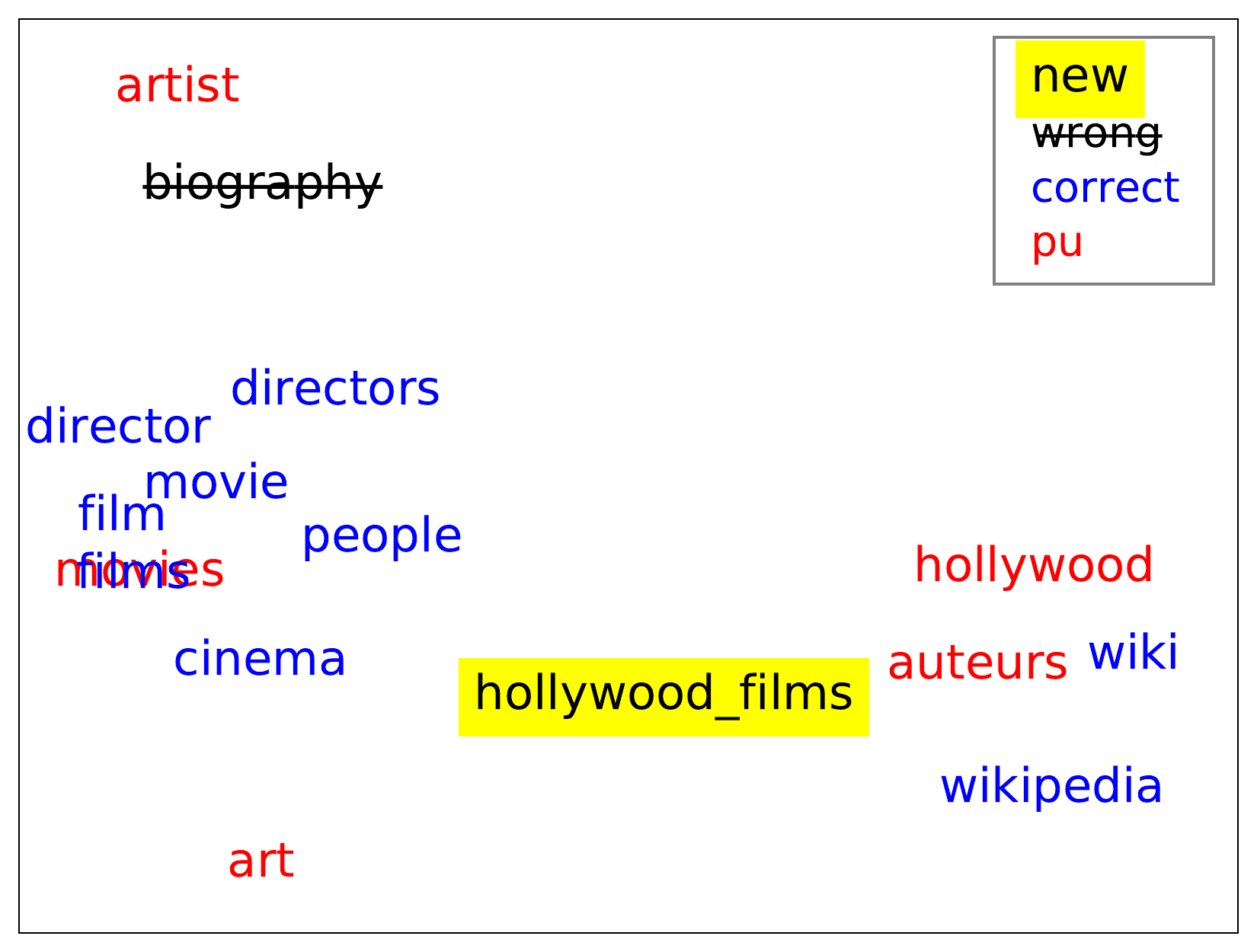}
}
\subfloat[\label{fig:ex_5} Brain Age]
{\includegraphics[trim=0 0 0 0,clip,width=0.33\linewidth]{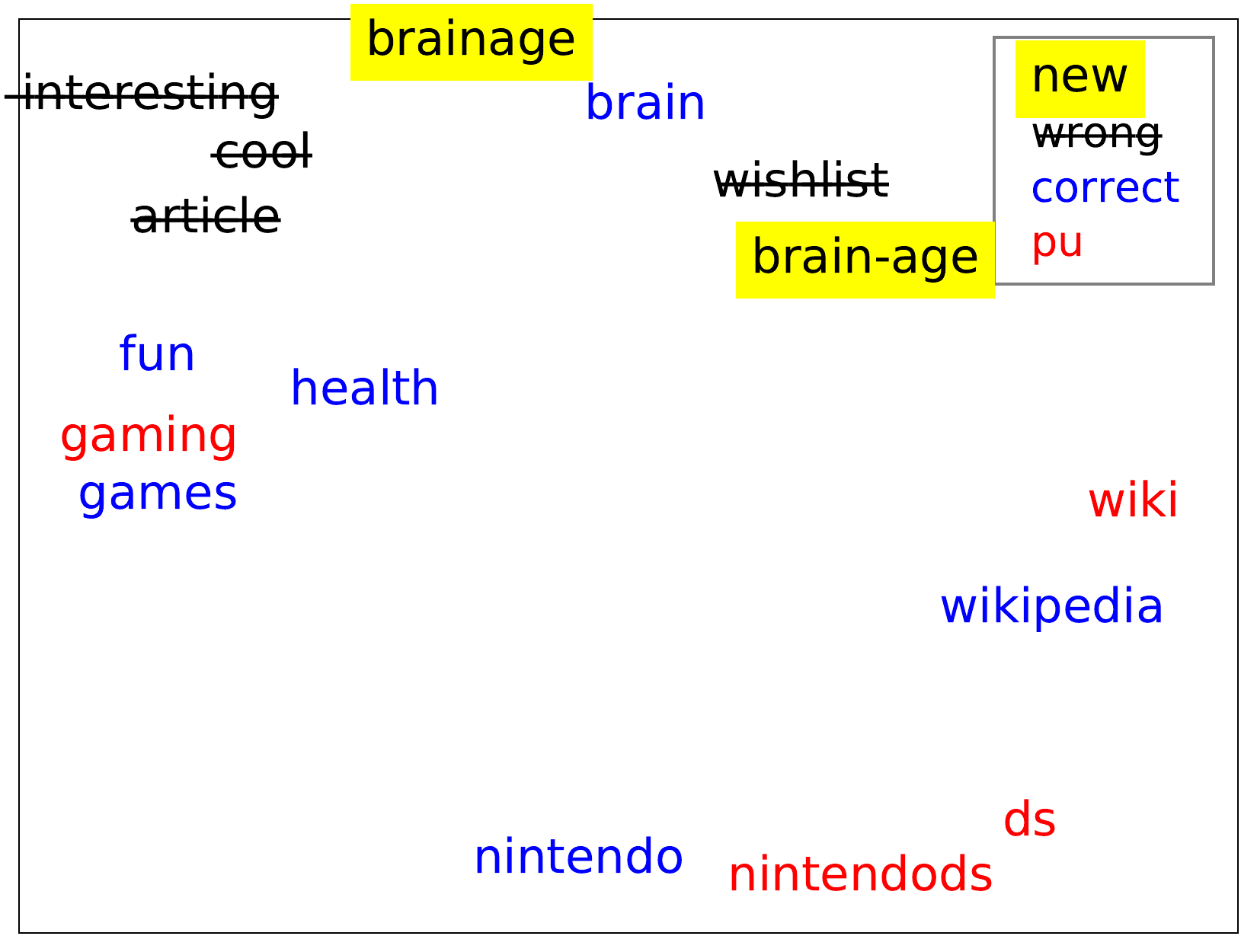}
}
\subfloat[\label{fig:ex_6} Croque-Monsieur]
{\includegraphics[trim=0 0 0 0,clip,width=0.33\linewidth]{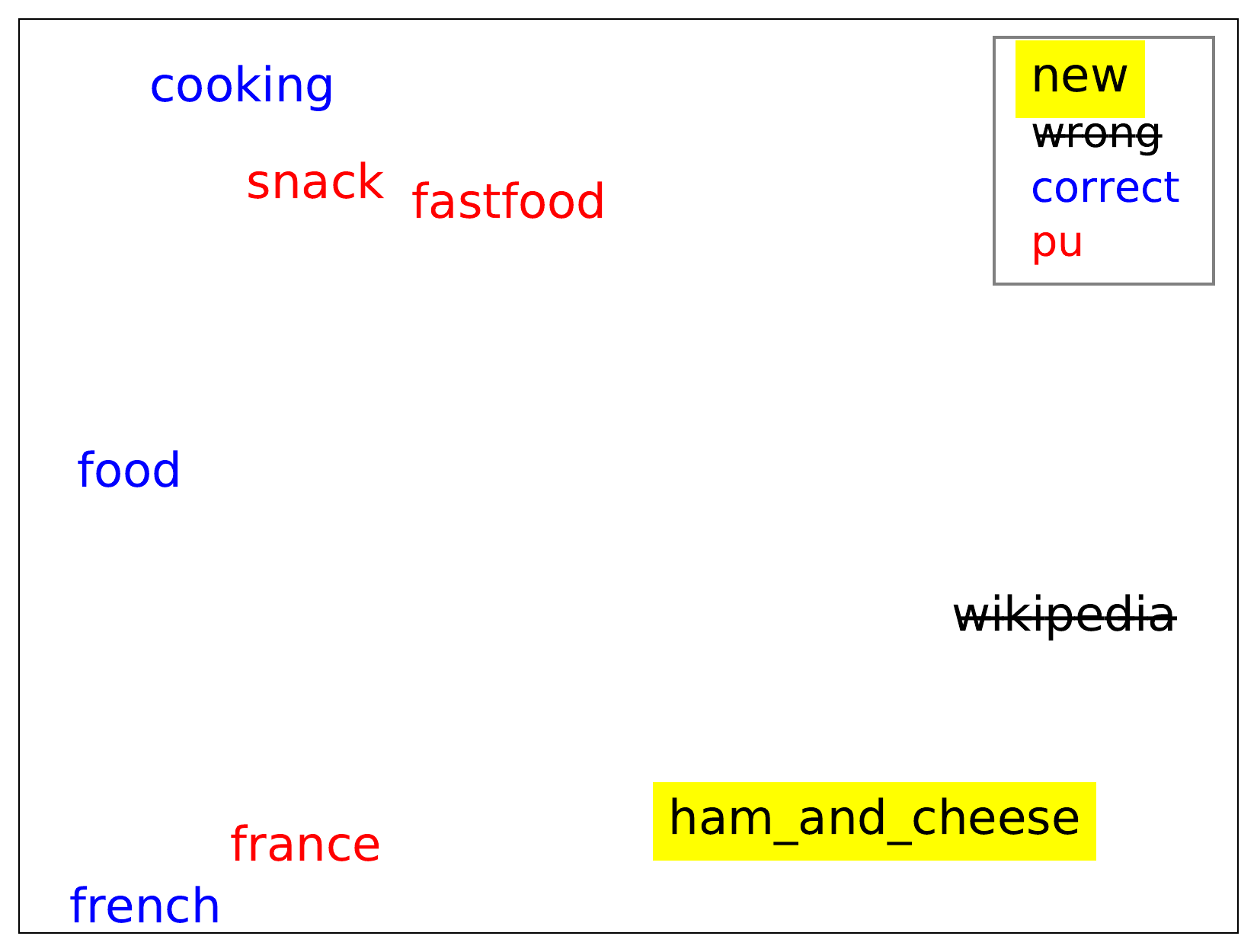}
}
\\
\subfloat[\label{fig:ex_7} Typography of Apple]
{\includegraphics[trim=0 0 0 0,clip,width=0.33\linewidth]{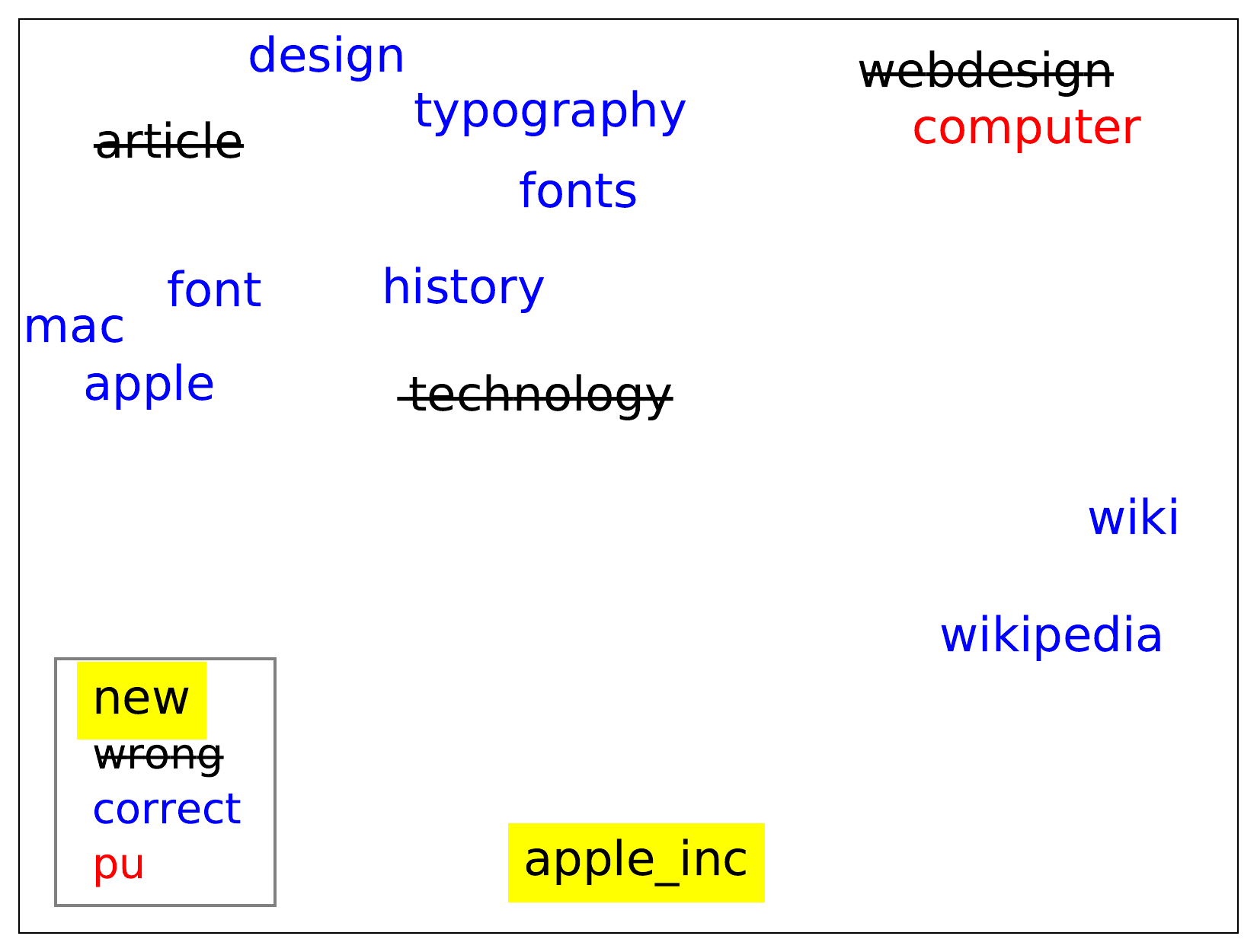}
}
\subfloat[\label{fig:ex_8} Jackalope]
{\includegraphics[trim=0 0 0 0,clip,width=0.33\linewidth]{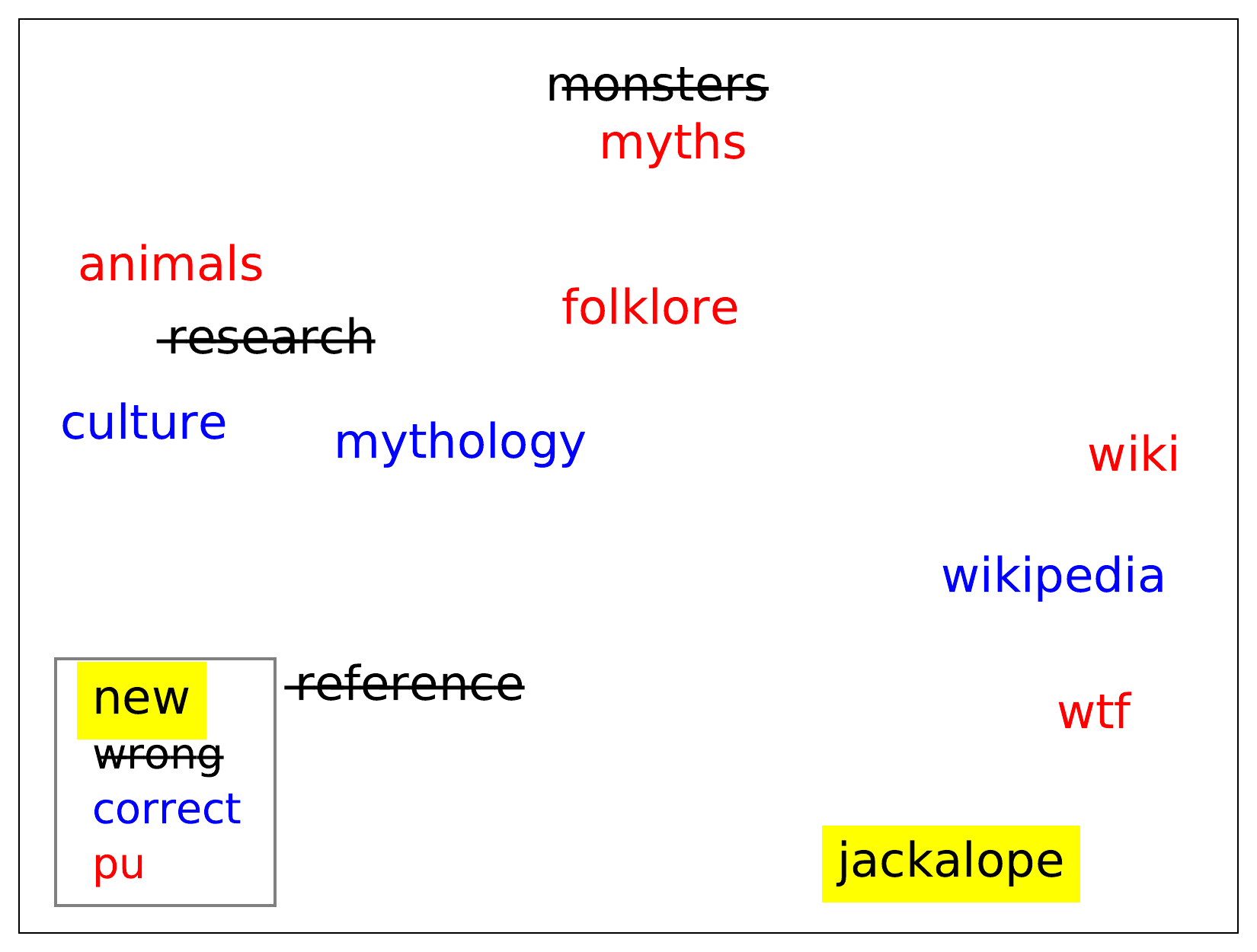}
}
\subfloat[\label{fig:ex_9} Anti-humor]
{\includegraphics[trim=0 0 0 0,clip,width=0.33\linewidth]{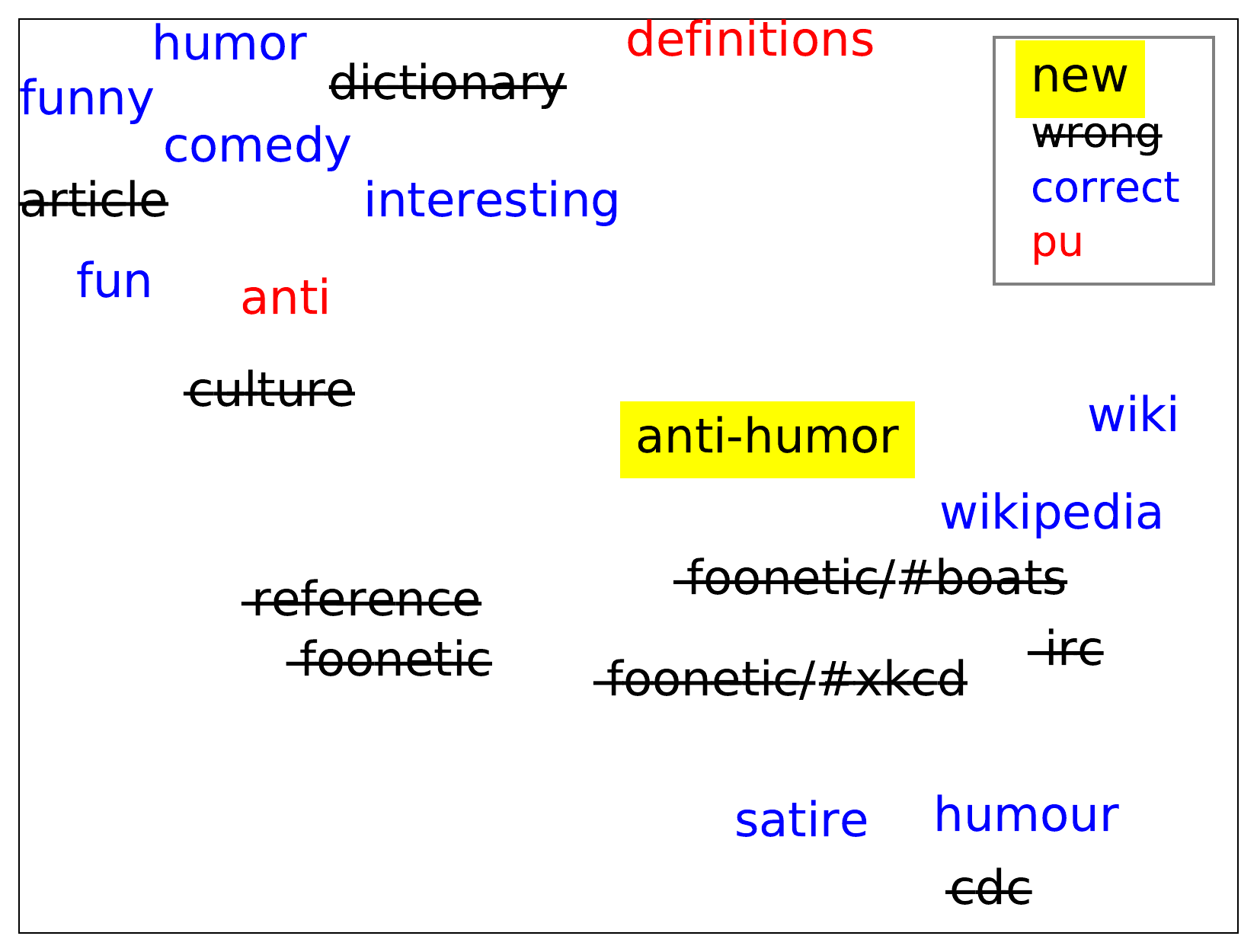}
}
\caption{\label{fig:label_semantics_appendix} Visualization of generated labels by \method-\texttt{BCL} for Wikipedia pages of annotated examples in Table~\ref{tab:quant_anal_appendix}.}
\end{figure*}

\subsection{Additional Analyses on Decoding Strategy}
\label{appendix:analysis_ensemble_generation}

\begin{table}[t!]
\center
\small{
\begin{tabular}{lcc|cc}
\hline
% \multirow{2}{*}{\method-\texttt{base}}
&\multicolumn{2}{c|}{\textsc{EurLex-4K}}
&\multicolumn{2}{c}{\textsc{Wiki10-31K}}\\
\cline{2-5}
& \textit{Mic.}& \textit{Mac.} & \textit{Mic.} & \textit{Mac.} \\
\hline
Greedy & 57.5&23.3&35.6&7.0\\
\hline
Beam (3) & \textbf{58.0} &23.7&35.7&7.8 \\
Beam (5) & \textbf{58.0} &\textbf{23.8}&\textbf{35.8}&\textbf{7.9}\\
Beam (10) & 57.9 &\textbf{23.8}&35.4&7.8\\
\hline
Tmp. (0.8) &53.7&21.9&30.6&7.1 \\
Top-$K$ (50) &51.7&20.8&28.7&6.8 \\ 
Top-$P$ (0.9) &53.2&21.1&28.8&6.5 \\ 
Top $P+K$ &53.6&21.5&31.1&7.4\\ 
\hline
\end{tabular}
\caption{Performances of \method-\texttt{base} trained with different decoding strategies. 
\vspace{-2mm}
}
\label{tab:performance_gen}
}
\end{table}

\begin{table}[ht!]
\center{
\small{
\begin{tabular}{lcc|cc}
\hline
\multirow{2}{*}{\method-\texttt{base}}&\multicolumn{2}{c|}{\textsc{EurLex-4K}}
&\multicolumn{2}{c}{\textsc{Wiki10-31K}}\\
\cline{2-5}
& \textbf{$Mic.$} & \textbf{$Mac.$}
& \textbf{$Mic.$} & \textbf{$Mac.$}\\
\hline
Beam (5) & \textbf{58.0} &23.8&\textbf{35.8}&7.9\\
\hline
Ens. Outer &53.5&\textbf{24.5} &31.5&\textbf{10.0} \\
Ens. Inner&54.0&18.9&29.2&4.2 \\
\hline
\end{tabular}
}
\caption{Task performance of \method-\texttt{base} from the best single stratagy (beam search with size 5) and ensemble generations on \textsc{EurLex-4K} and \textsc{Wiki10-31K}. 
The highest scores are \textbf{bold}.
\vspace{-3mm}}
\label{tab:performance_ens}
}
\end{table}

Followed by Figure~\ref{subfig:ablation_decoder}, Table~\ref{tab:performance_gen} shows a task performance across various decoding strategies, including different beam size for beam search and a single sampling restriction.

Additionally, instead of choosing single generation strategy, we can even consider to integrate generation outputs from different generation strategies.
For ensemble generations, we choose three single generation strategies; beam search with size 5, Top $K+P$ sampling and sampling with temperature 0.8 to get diverse label sequences.
We also consider two different types of joining method; inner join to union all labels and outer join to intersect labels from single generations.

Table~\ref{tab:performance_ens} shows task performance of ensemble generations. 
We find that outer joining ensemble generation could improve macro F1 scores as it includes more labels than single result.
However, it simultaneously drops other micro F1 scores due to the high chance to contain wrongly predicted labels as well.
On the other hand, inner joining ensemble generation in general harms the performance by restricting predicted labels occurring at any single generations, though this yields higher micro F1 scores than inner joining ensemble results.

\subsection{Additional Analyses on Clustering and Representation}
\label{appendix:analysis_clustering_representation}

\begin{table}[ht!]
\center
\small{
\begin{tabular}{lcc|cc}
\hline
\multirow{2}{*}{\method-\texttt{base}}&\multicolumn{2}{c|}{\textsc{EurLex-4K}}
&\multicolumn{2}{c}{\textsc{Wiki10-31K}}\\
\cline{2-5}
& \textbf{$Mic.$} & \textbf{$Mac.$}
& \textbf{$Mic.$} & \textbf{$Mac.$}\\
\hline
Kmn. + tf-idf & 58.4 &24.1 &36.6&\textbf{8.9} \\
Kmn. + t5-enc. & \textbf{58.5} &23.9&36.8&8.8  \\ 
\hline
Ahcl. + tf-idf &58.4&\textbf{24.4}&36.8&8.8\\ 
Ahcl. + t5-enc &58.0&24.0&\textbf{37.0}&8.6\\
\hline
\end{tabular}
}
\caption{Task performances trained with different cluster algorithm and input features on \textsc{EurLex-4K} and \textsc{Wiki10-31K}.
Here we fix cluster size as 30.
The highest scores are \textbf{bold}.
\vspace{-3mm}}
\label{tab:performance_cluster}
\end{table}

% We first compare options that affect a cluster quality such as algorithm and features to find an optimal choice for cluster-based \method architectures.
We compare two clustering algorithms; K-means and Agglomerative hierarchical clustering, and two text representations for the label features; TF-IDF and the last hidden states of T5 encoder in Table~\ref{tab:performance_cluster}.
We find that both algorithms show comparable performances.
As computing cost is more expensive in Agglomerative hierarchical clustering, we mainly use K-means in our experiments. %except \textsc{EUR-{EUR-EUR-UR-R--Lex}.
For text representation, the pre-trained T5 encoder achieves similar or slightly better performance to TF-IDF vectors.
Pre-trained T5 encoder is more efficient in training as it has much lower size of dimensionality (e.g., 768 in t5-base) than tfidf (e.g., >100,000 for both \textsc{EUR-Lex} and \textsc{Wiki10-31K}).
Thus, for all experiments with clustering method, we use K-means with pre-trained T5 encoder text representation.

\subsection{Additional Examples of Human Annotation}
\label{appendix:human_annotation}
In Table~\ref{tab:quant_anal_appendix}, We provide more annotation examples from \textsc{Wiki10-31K}, following the Table~\ref{tab:quant_anal} to show how \method generates labels.
In Figure~\ref{fig:label_semantics_appendix}, we also provide visualizations of generated labels by \method for examples in Table~\ref{tab:quant_anal_appendix}.

\begin{table*}[t!]
\centering
\small
{
\begin{adjustbox}{width=0.94\linewidth,center}
\begin{tabularx}{\textwidth}{@{}m{6.0cm}m{1.7cm}m{7.2cm}@{}}
\hline
\textbf{Input Document} & \textbf{Models} & \textbf{Labels} \\
\hline
\multirow{3}{\linewidth}{\setlength{\fboxsep}{0pt}\colorbox{yellow!90}{Emily Elizabeth Dickinson} (December 10, 1830– May 15, 1886) was an American poet.  Born in Amherst, Massachusetts to a successful family with strong community ties, she lived a mostly introverted and reclusive life. 
After she studied at the Amherst Academy for seven years in her youth, she spent a short time at ...} & True& 
authors biography dickinson emily journal library literature openaccess people poem poet poetry reference research to-read wiki wikipedia writers\\
\cdashline{2-3}
&AttentionXML & 
\textcolor{blue}{wiki poet writers wikipedia literature authors} 
\textcolor{red}{books} \sout{writing} \sout{history} \textcolor{red}{poets writer} \textcolor{blue}{people poetry biography} \sout{inspiration} \textcolor{blue}{american poems} \sout{luule} \\
\cdashline{2-3}
& \method-\texttt{BCL} &
\textcolor{blue}{wikipedia wiki people} \sout{art} \textcolor{red}{books} \textcolor{blue}{literature}
\sout{english}
\textcolor{blue}{poetry writers} 
\textcolor{red}{writer} 
\textcolor{blue}{poet} \textcolor{red}{elizabeth} \textcolor{blue}{dickinson} \setlength{\fboxsep}{0pt}\colorbox{yellow!90}{emilydickinson}\\
\hline
\multirow{3}{\linewidth}{
Screenshot of vimeo.com home page \setlength{\fboxsep}{0pt}\colorbox{yellow!90}{Vimeo} is a video-centric social network site (owned by IAC/InterActiveCorp) which launched in November 2004. The site supports embedding, sharing, video storage, and allows user-commenting on each video page...} & True& articles computer reference socialnetworks technology tools video web2.0 wikipedia\\
\cdashline{2-3}
&AttentionXML & 
\textcolor{blue}{video web2.0 wikipedia}  \textcolor{red}{wiki} \sout{media} \sout{youtube} \textcolor{red}{videos} \sout{videoblogging} \sout{streaming}\\
\cdashline{2-3}
& \method-\texttt{BCL} &
\textcolor{blue}{wikipedia} \textcolor{red}{wiki} \textcolor{blue}{reference technology} \textcolor{red}{web internet} \sout{social} \textcolor{blue}{video web2.0} 
\sout{no\_tag} \textcolor{red}{socialnetworking} \sout{socialsoftware} \sout{phd} %\textcolor{red}{socialnet} 
\textcolor{red}{social\_networking social\_network} 
%\sout{ian\_minimalism} 
\setlength{\fboxsep}{0pt}\colorbox{yellow!90}{vimeo}
\\
\hline
\multirow{3}{\linewidth}{Diet Coke and Mentos Eruption is a reaction of Diet Coke and mint Mentos candies, a bottle of Diet Coke (other \setlength{\fboxsep}{0pt}\colorbox{yellow!90}{carbonated beverages} may be used instead) and dropping some Mentos. This causes the Coke to foam at a rapid rate and spew into the air... } & True&beverage candy chemistry coca-cola coke dietcoke drink eruption experiment experiments video explosion food fun funny interesting mint prank science \\
\cdashline{2-3}
&AttentionXML & 
\textcolor{red}{wikipedia} \textcolor{blue}{fun science} \sout{diet} \textcolor{red}{wiki} \textcolor{blue}{funny coke} \sout{tv} \textcolor{blue}{video} 
\sout{health} \textcolor{blue}{interesting humor food} \\
\cdashline{2-3}
& \method-\texttt{BCL} &
\textcolor{red}{wikipedia wiki} \textcolor{blue}{science interesting fun video funny food} \textcolor{red}{humor weird humour wtf} \sout{\#afterdarkclub} \setlength{\fboxsep}{0pt}\colorbox{yellow!90}{soda} \textcolor{blue}{eruption}\\
\hline
\multirow{3}{\linewidth}{David Leo Fincher (born August 28, 1962) is an Academy Award-nominated \setlength{\fboxsep}{0pt}\colorbox{yellow!90}{American filmmaker} and music video director known for his dark and stylish movies such as Seven, Fight Club, Zodiac and The Curious Case of Benjamin Button...} & True& cinema david director directors figures film filmmaking films fincher inspiration movie people wiki wikipedia\\
\cdashline{2-3}
&AttentionXML & 
\textcolor{blue}{wiki directors} \textcolor{red}{video} \textcolor{blue}{wikipedia cinema}  \sout{pitt} \textcolor{blue}{people director films} \textcolor{red}{movies} \textcolor{blue}{movie film} \textcolor{red}{filmmaker} \sout{brad}  \\
\cdashline{2-3}
& \method-\texttt{BCL} &
\textcolor{blue}{wikipedia wiki people} \textcolor{red}{art} \textcolor{blue}{film} \sout{biography} \textcolor{red}{movies artist} \textcolor{blue}{movie cinema films director directors}
\textcolor{red}{auteurs hollywood}
\setlength{\fboxsep}{0pt}\colorbox{yellow!90}{hollywood\_films} \\
\hline
\multirow{4}{\linewidth}{\setlength{\fboxsep}{0pt}\colorbox{yellow!90}{Brain Age}: Train Your Brain in Minutes a Day!, also known as Dr. Kawashima's Brain Training: How Old Is Your Brain? in PAL regions, is an entertainment video game that employs puzzles. It was developed and published by the video gaming company Nintendo for the Nintendo DS handheld video game console...} & True& @mentat biology brain braintraining computer exercise fitness fun game games health medical nintendo read science sudoku unit4 wikipedia\\
\cdashline{2-3}
&AttentionXML & 
\textcolor{blue}{wikipedia game games fun science nintendo sudoku brain} \sout{mind} \textcolor{red}{ds} \textcolor{red}{wiki} \sout{video} \sout{memory} \textcolor{red}{gaming puzzle puzzles} \textcolor{red}{videogames} \textcolor{red}{nds}  \\
\cdashline{2-3}
& \multirow{2}{\linewidth}{\method-\texttt{BCL}} &
\textcolor{blue}{games wikipedia fun health brain nintendo}  
\textcolor{red}{nintendods wiki gaming} 
\sout{wishlist} \sout{article} \sout{interesting} \sout{cool} \textcolor{red}{ds} 
\setlength{\fboxsep}{0pt}\colorbox{yellow!90}{brain-age} \setlength{\fboxsep}{0pt}\colorbox{yellow!90}{brainage}\\
\hline
\multirow{4}{\linewidth}{A croque-monsieur is a hot \setlength{\fboxsep}{0pt}\colorbox{yellow!90}{ham and cheese} (typically emmental[citation needed] or gruyère) grilled sandwich. It originated in France as a fast-food snack served in cafés and bars ...} & True& cooking food french recipe sandwich\\
\cdashline{2-3}
&AttentionXML & 
\textcolor{blue}{food}
\sout{wikipedia}
\textcolor{blue}{cooking french}
\sout{wiki}\\
\cdashline{2-3}
& \multirow{2}{\linewidth}{\method-\texttt{BCL}} & \sout{wikipedia} \textcolor{blue}{food}  \textcolor{red}{france} \textcolor{blue}{french cooking} 
\setlength{\fboxsep}{0pt}\colorbox{yellow!90}{ham\_and\_cheese}  
\textcolor{red}{fastfood snack}
\\
\hline
\multirow{4}{\linewidth}{Typography of \setlength{\fboxsep}{0pt}\colorbox{yellow!90}{Apple Inc.} refers to Apple Inc.’s use of typefaces in marketing, operating systems, and industrial design. Apple has used three corporate fonts throughout its history: Motter Tektura, Apple Garamond and Adobe Myriad.  For at least 18 years, Apple's corporate font was a custom variant of the ITC Garamond typeface, called Apple Garamond ...} & True& adobe apple branding chronology computer computers design design.fonts fmp font fonts helpful history imac ipod list mac macintosh marketing myriad print pro product reference sda spunti storia typography wiki wikipedia\\
\cdashline{2-3}
&AttentionXML & 
\textcolor{blue}{typography fonts apple font design wikipedia} \textcolor{red}{type typeface} \textcolor{blue}{wiki} \textcolor{red}{tipografia} \sout{ttf} \textcolor{blue}{macintosh reference mac history} graphics \sout{logo} graphic \sout{designers webdesign graphicdesign diseño} \textcolor{red}{computer typographer} \sout{ipod} \sout{brand} \sout{article} \sout{technology} \sout{business} \sout{advertising} \\
\cdashline{2-3}
& \multirow{2}{\linewidth}{\method-\texttt{BCL}} & 
\textcolor{blue}{wikipedia wiki history} 
\sout{article} 
\textcolor{blue}{design}
\sout{technology}
\textcolor{red}{computer}
\sout{webdesign}
\textcolor{blue}{mac apple typography fonts font}  
\setlength{\fboxsep}{0pt}\colorbox{yellow!90}{apple\_inc}
\\
\hline
\multirow{4}{\linewidth}{\setlength{\fboxsep}{0pt}\colorbox{yellow!90}{The jackalope} — also called an antelabbit, aunt benny, Wyoming thistled hare or stagbunny — is an imaginary animal of folklore and a supposed cross between a jackrabbit and an antelope, goat, or deer, which is usually ...} & True& american animal creatureproject cryptozoology culture fiction humor humour myth mythology storyideas wikipedia \\
\cdashline{2-3}
&AttentionXML & 
\textcolor{red}{folklore} \textcolor{blue}{wikipedia} \textcolor{red}{animals} \textcolor{blue}{mythology} \textcolor{red}{wiki} \textcolor{blue}{cryptozoology culture} \sout{monsters} \textcolor{blue}{animal} \textcolor{red}{weird interesting} \textcolor{blue}{myth}
\\
\cdashline{2-3}
& \multirow{2}{\linewidth}{\method-\texttt{BCL}} & 
\textcolor{blue}{wikipedia} \textcolor{red}{wiki} \sout{reference} \sout{research} \textcolor{blue}{culture mythology} \textcolor{red}{animals folklore wtf myths} \sout{monsters} \setlength{\fboxsep}{0pt}\colorbox{yellow!90}{jackalope}
\\
\hline
\multirow{4}{\linewidth}{\setlength{\fboxsep}{0pt}\colorbox{yellow!90}{Anti-humor and anti-jokes}[1] (also known as unjokes) are a kind of humor based on the surprise factor of absence of an expected joke or of a punch line in a narration which is set up as a joke.
 This kind of anticlimax is similar to that of the shaggy dog story.[2] In fact, John Henderson sees the "shaggy dog story" ... } & True& comedy favourites fun funny humor humour information interesting jokes people postmodernism wiki wikipedia \\
\cdashline{2-3}
&AttentionXML & 
\textcolor{blue}{humor wikipedia funny comedy humour fun} 
\textcolor{red}{satire} 
\textcolor{blue}{wiki} 
\sout{dog} \sout{animals} \sout{standup} \sout{parody} \textcolor{red}{joke}
\\
\cdashline{2-3}
& \multirow{2}{\linewidth}{\method-\texttt{BCL}} & 
\textcolor{blue}{wikipedia wiki} \sout{reference} \textcolor{blue}{interesting} \sout{article} \sout{culture} \textcolor{blue}{fun funny humor} \sout{irc} \sout{foonetic} \sout{foonetic/\#xkcd} \textcolor{red}{definitions} \sout{dictionary} \textcolor{blue}{humour comedy} \textcolor{red}{satire} \textcolor{red}{anti} \sout{cdc} \sout{foonetic/\#boats} \setlength{\fboxsep}{0pt}\colorbox{yellow!90}{anti-humor}
\\
\hline
\end{tabularx}
\end{adjustbox}
}
\caption{Additional examples of ground-truth and predicted labels in \textsc{Wiki10-31K}, following Table~\ref{tab:quant_anal}. 
\vspace{-3mm}
}
\label{tab:quant_anal_appendix}
\end{table*}

\end{document}